\documentclass[10pt, a4paper]{article}
\usepackage[final]{lrec2026}

\usepackage{algorithm}
\usepackage{algorithmic}
\usepackage{arydshln}
\usepackage{multirow}
\usepackage{multicol}
\usepackage[T5]{fontenc}
\usepackage{subcaption}
\usepackage{amsmath}
\usepackage{tabularray}
\usepackage{xcolor}
\usepackage{longtable}
\usepackage{pifont}
\usepackage{natbib}  
\usepackage{caption}
\usepackage{graphicx}
\usepackage{newfloat}
\usepackage{listings}
\usepackage{array}

\title{ViX-Ray: A Vietnamese Chest X-Ray Dataset for Vision-Language Models}

\name{
\begin{tabular}{c}
Duy Vu Minh Nguyen\textsuperscript{1,2}, Chinh Thanh Truong\textsuperscript{2}, Phuc Hoang Tran\textsuperscript{3} \\
Hung Tuan Le\textsuperscript{4, 5}, Nguyen Van-Thanh Dat\textsuperscript{4, 5} \\
Trung Hieu Pham\textsuperscript{3}, Kiet Van Nguyen\textsuperscript{4, 5, *}
\end{tabular}
}

\address{Industrial University of Ho Chi Minh City\textsuperscript{1}, Military Hospital 175\textsuperscript{2}, Pythera AI\textsuperscript{3} \\ University of Information Technology, Ho Chi Minh City, Vietnam\textsuperscript{4}\\Vietnam National University, Ho Chi Minh City, Vietnam\textsuperscript{5} \\
         duynvm@\{iuh.edu.vn, benhvien175.vn\}\\
         chinhtt@benhvien175.vn, phuc.tran@pythera.ai\\
         \{21520250, 20520436\}@gm.uit.edu.vn\\
         hieu.pham@pythera.ai, kietnv@uit.edu.vn}
\abstract{
Vietnamese medical research has become an increasingly vital domain, particularly with the rise of intelligent technologies aimed at reducing time and resource burdens in clinical diagnosis. Recent advances in vision-language models (VLMs), such as Gemini and GPT-4V, have sparked a growing interest in applying AI to healthcare. However, most existing VLMs lack exposure to Vietnamese medical data, limiting their ability to generate accurate and contextually appropriate diagnostic outputs for Vietnamese patients. To address this challenge, we introduce ViX-Ray, a novel dataset comprising 5,400 Vietnamese chest X-ray images annotated with expert-written findings and impressions from physicians at a major Vietnamese hospital. We analyze linguistic patterns within the dataset, including the frequency of mentioned body parts and diagnoses, to identify domain-specific linguistic characteristics of Vietnamese radiology reports. Furthermore, we fine-tune five state-of-the-art open-source VLMs on ViX-Ray and compare their performance to leading proprietary models, GPT-4V and Gemini. Our results show that while several models generate outputs partially aligned with clinical ground truths, they often suffer from low precision and excessive hallucination, especially in impression generation. These findings not only demonstrate the complexity and challenge of our dataset but also establish ViX-Ray as a valuable benchmark for evaluating and advancing vision-language models in the Vietnamese clinical domain.
\\ \newline \Keywords{Vietnamese X-Ray Caption, Medical Multimodal Learning, Vision Language Model Medical} }

\begin{document}

\maketitleabstract
\section{Introduction}

\renewcommand{\thefootnote}{\fnsymbol{footnote}}
\footnotetext[1]{Corresponding author.}
\renewcommand{\thefootnote}{\arabic{footnote}}

\begin{figure}[htb]
    \centering
    \includegraphics[width=0.8\linewidth]{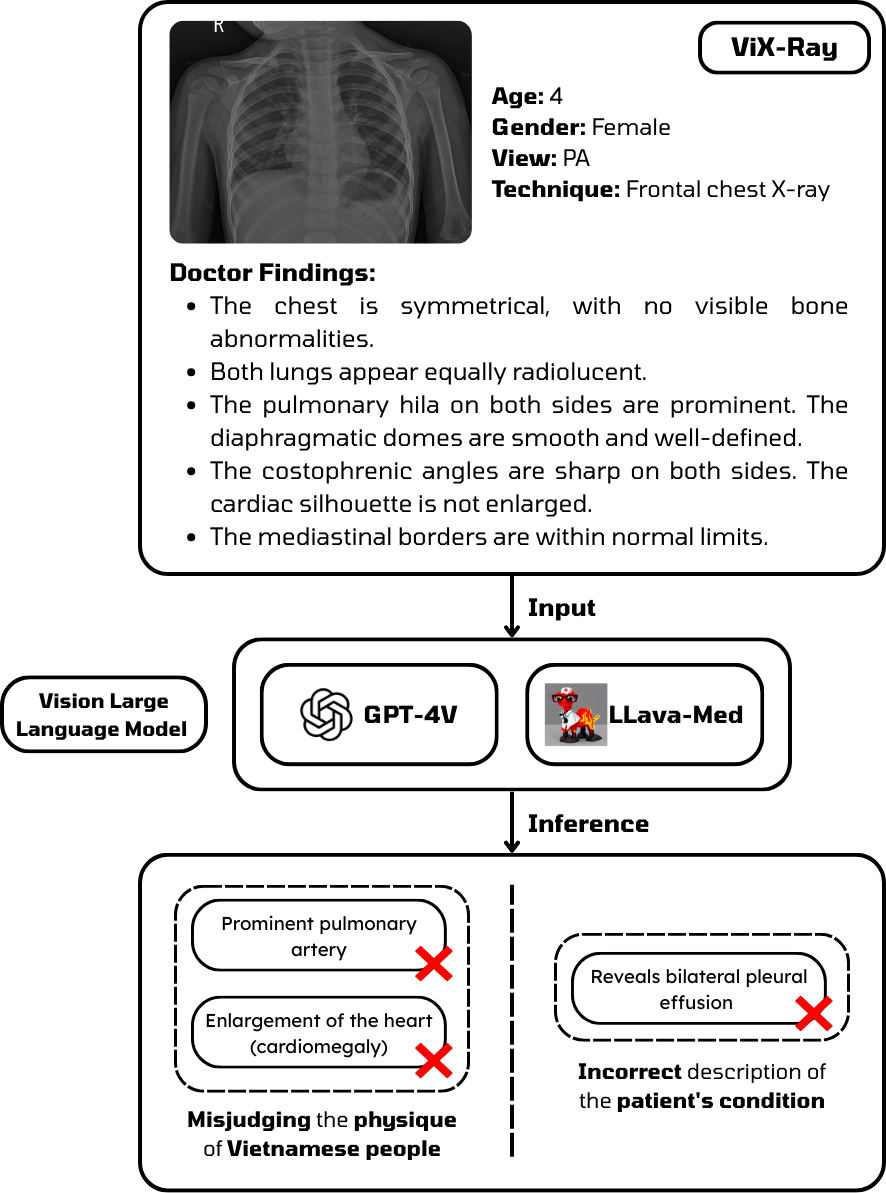}
    \caption{An illustrative example of misdiagnosing the condition of a Vietnamese patient using English Vision Language Models.}
    \label{fig:motivation}
\end{figure}

\begin{table*}[htb]
\centering
\resizebox{\textwidth}{!}{
\begin{tabular}{lccccc}
\hline
\multicolumn{1}{c}{\textbf{Dataset}} & \textbf{Year} & \textbf{Size} & \textbf{Text Findings} & \textbf{Text Impressions} & \textbf{Vietnamese Image} \\ \hline
Montgomery County chest X-ray & 2014 & 138 & \textcolor{blue}{\ding{51}} & \textcolor{red}{\ding{55}} & \textcolor{red}{\ding{55}} \\ 
ChestX-ray8 & 2017 & 108,948 & \textcolor{blue}{\ding{51}} & \textcolor{blue}{\ding{51}} & \textcolor{red}{\ding{55}} \\ 
Padchest & 2019 & 160,868 & \textcolor{blue}{\ding{51}} & \textcolor{blue}{\ding{51}} & \textcolor{red}{\ding{55}} \\ 
CheXpert & 2019 & 224,316 & \textcolor{blue}{\ding{51}} & \textcolor{blue}{\ding{51}} & \textcolor{red}{\ding{55}} \\ 
VinDr-RibCXR & 2021 & 245 & \textcolor{red}{\ding{55}} & \textcolor{red}{\ding{55}} & \textcolor{blue}{\ding{51}} \\ 
VinDr-CXR & 2022 & 18,000 & \textcolor{red}{\ding{55}} & \textcolor{red}{\ding{55}} & \textcolor{blue}{\ding{51}} \\ 
PediCXR & 2023 & 9,125 & \textcolor{red}{\ding{55}} & \textcolor{red}{\ding{55}} & \textcolor{blue}{\ding{51}} \\ \hdashline
\textbf{ViX-Ray (Ours)} & 2025 & 5,400 & \textcolor{blue}{\ding{51}} & \textcolor{blue}{\ding{51}} & \textcolor{blue}{\ding{51}} \\ \hline
\end{tabular}
}
\caption{Summary of public chest radiographic datasets with metadata. "Vietnamese Image" denotes whether the data set includes chest radiographs of Vietnamese subjects.}
\label{tab:dataSurvey}
\end{table*}

Clinical X-ray research is one of the most prominent areas in the medical field, aiming to extract valuable insights from X-ray images, such as identifying damaged organs, assessing patient conditions, and more. Consequently, large-scale datasets such as CheXpert \citep{chexpert}, ChestX-ray8 \citep{Chestx-ray8}, and ChestX-ray14 \citep{Chestx-ray8} have been introduced. These datasets have enabled the development of high-performing models that can address real-world problems based on X-ray images of patients \citep{ResNet, medvit,medclip}. In recent years, the emergence of Vision-Language Models (VLMs) such as LLaVA-Med \citep{llava-med} and GPT-4V \citep{gpt-4_technical_report,gpt-4v} has further advanced the field \citep{gpt-4v}. These models can interpret X-ray images, describe patient characteristics, and generate preliminary diagnoses, offering substantial value in practical medical scenarios. However, most of the publicly available datasets have been collected in Western countries \citep{chexpert, Chestx-ray8}, where epidemiological profiles, physiological characteristics, lifestyle habits, and environmental factors differ significantly from those of the Vietnamese population \citep{nickol1982radiographic, donnelly1991factors, bild2005ethnic}. Consequently, models trained on these datasets often exhibit limited generalizability and may not perform well when applied to Vietnamese patients \citep{glocker2023risk}. For instance, in the case illustrated in Figure \ref{fig:motivation}, two state-of-the-art VLMs failed to accurately describe the condition of a Vietnamese patient, highlighting the critical need for a dedicated Vietnamese X-ray dataset with detailed annotations.

In recent years, the Vietnamese medical AI community has made commendable efforts to develop large-scale X-ray datasets such as VinDr-CXR \citep{VinDr-CXR}, VinDr-Mammo \citep{VinDr-Mammo}, and VinDr-RibCXR \citep{VinDr-RibCXR}, primarily targeting tasks like image classification and segmentation. Additionally, datasets like ViNewsQA \citep{ViNewsQA} and ViMedAQA \citep{ViMedAQA} have been introduced to support Vietnamese medical question answering. However, these datasets still exhibit certain limitations. Most image-based datasets are confined to tasks such as disease classification or rib segmentation \citep{VinDr-RibCXR}, thereby restricting the range of applicable tasks. Meanwhile, the medical QA datasets often lack detailed clinical information, including lesion descriptions or diagnostic conclusions from medical experts \citep{nguyen2019overcoming, oakden2020exploring}. As a result, the answers generated tend to be general rather than clinically insightful. These limitations emphasize the urgent need for a comprehensive Vietnamese X-ray dataset enriched with detailed patient information and expert-level annotations and diagnoses specifically tailored to the Vietnamese population.

Motivated by the aforementioned challenges, this paper introduces a new dataset consisting of 5,400 samples. Each sample includes a chest X-ray image, anonymized administrative information, and pathological descriptions written by certified radiologists. The data were collected from patients who underwent examinations at a hospital in Vietnam, and the study received ethical approval from the institutional review board of the hospital.


We conduct statistical analyses on key characteristics of the ViX-Ray dataset, such as diagnosis frequency and body part frequency, to highlight the linguistic patterns found in the medical reports. For the experimental setup, we evaluate a diverse set of Vision-Language Models (VLMs), including Vietnamese-specific models like Vintern-1B-v3.5 \citep{vintern} and Lavy \citep{lavy}, as well as multilingual models trained with Vietnamese data, such as InternVL2.5 \citep{internvl-2.5}, Qwen2.5-VL \citep{qwen2.5-vl}, and MiniCPM-V-2.6 \citep{minicpm-v}, covering model sizes ranging from 2B to 7B parameters, with and without instruction tuning.

Given that the dataset includes both descriptive and diagnostic annotations written by medical experts, we propose a three-stage evaluation pipeline. In the first stage, models are prompted to describe the condition of the patient using only the chest X-ray image. In the second stage, models are asked to diagnose based on the same input. The third stage involves a multi-turn interaction, in which models are required to first describe the condition and then offer a diagnosis through a subsequent conversational turn. At each stage, model performance before and after supervised fine-tuning (SFT) is compared, offering a detailed analysis of the impact of fine-tuning on effectiveness in the Vietnamese medical context.

Our experimental results show that Qwen2.5-VL-7B achieves the best overall performance across all stages of the evaluation pipeline. We further compare its performance with two leading proprietary models, GPT-4V (o4 multimodal version) \citep{gpt-4o-v} and Gemini \citep{gemini}, demonstrating its superior diagnostic precision and practical potential to support real-world clinical workflows and alleviate the burden on healthcare professionals. In addition, we publicly release our dataset on Hugging Face\footnote{datasets$/$MilitaryHospital175$/$VNMedical\_bv175} to support the research community and encourage further studies.

\section{Related Work}

In the global context, medical research in general, and chest X-ray research in particular, has a long-standing history with the development of numerous diverse datasets. These range from small-scale datasets such as the Montgomery County Chest X-ray dataset \citep{jaeger2014two} with 138 frontal chest X-rays, and the Shenzhen Chest X-ray dataset \citep{jaeger2014two} with 662 frontal images, to larger-scale collections such as ChestX-ray8 \citep{Chestx-ray8} with 108,948 frontal X-ray images, and its expanded version ChestX-ray14 with 112,120 X-ray images. Other notable examples include PadChest \citep{Padchest}, which contains 160,868 images obtained from more than 67,000 patients, and MIMIC-CXR \citep{MIMIC-CXR}, which features 377,110 chest radiographs with frontal and lateral views. Alongside these datasets, the research community has explored a wide range of downstream tasks such as pneumonia detection \citep{rajpurkar2017chexnet, zhang2023biomedclip}, medical image generation \citep{gibson2018niftynet,welander2018generative}, and thoracic disease classification \citep{ranjan2018jointly, zunaed2024learning, ashraf2023synthensemble}. Furthermore, these data sets have paved the way for multimodal research, exemplified by data sets such as RadVisDial \citep{kovaleva-etal-2020-towards}, which utilizes X-ray images from MIMIC-CXR \citep{MIMIC-CXR}, and SLAKE \citep{liu2021slake}, which aggregates images from various sources \citep{Slake_cite, Chestx-ray8}. These resources have significantly advanced studies in Medical Visual Question Answering (VQA) \citep{llava-med, pubmedclip}.

Several efforts in the past five years have focused on developing medical datasets, especially for chest X-ray tasks. For example, VinDr-CXR \citep{VinDr-CXR} contains 18,000 annotated images selected from 100,000 chest radiographs, labeled by 17 experienced radiologists. VinDr-RibCXR \citep{VinDr-RibCXR} targets rib segmentation and labeling. For pediatric patients, \citeauthor{pedicxr} introduced PediCXR, which includes 9,125 posterior-anterior chest radiographs of children under 10 to support research on thoracic disease detection and classification. These datasets were collected from reputable hospitals in Vietnam, such as Hospital 108 and Hanoi Medical University Hospital, with ethical approvals. Since the images come from Vietnamese patients, they offer a valuable, population-specific resource reflecting local physical and medical characteristics. However, most focus mainly on classification or segmentation tasks \citep{VinDr-RibCXR, pedicxr, VinDr-CXR}, lacking detailed descriptions or clinical diagnoses. In contrast, our dataset (see Table \ref{tab:dataSurvey}) includes expert-level findings and diagnostic conclusions certified radiologists make, offering rich annotations to support broader research and practical applications for the Vietnamese population.

\section{Dataset}
\begin{table*}[t]
    \centering
    
    \begin{tabular}{:c:p{0.9\linewidth}:}
    \hdashline
        \textbf{Age} &  27\\
        \textbf{Gender} & Nam - (\textit{Male})\\
        \textbf{View} &  PA (\textit{Posterior Anterior})\\
        \textbf{Note} &  Chụp Xquang ngực thẳng - (\textit{Frontal chest X-ray}\\
        \textbf{Technique} &  Chụp Xquang tim phổi thẳng - (\textit{Posteroanterior chest radiograph)}\\ \hdashline
        \multirow{7}{*}{\textbf{Findings}}& Hình ảnh gãy cung xương sườn 2 bên trái. Phổi hai bên kém sáng đều. Thâm nhiễm nhu mô phổi hai bên, đám mờ đông đặc nhu mô phổi phải. Bờ vòm hoành hai bên đều. Góc sườn hoành hai bên nhọn. Bóng tim không to. Các bờ trung thất trong giới hạn bình thường. \\
        & (\textit{Fracture of the left 2nd rib arch. Both lungs show decreased translucency. Bilateral pulmonary parenchymal infiltrates, with a consolidation opacity in the right lung parenchyma. Both diaphragmatic domes have clear borders. Both costophrenic angles are sharp. The cardiac silhouette is not enlarged. Mediastinal borders are within normal limits.})\\ \hdashline
        \multirow{4}{*}{\textbf{Impressions}} & Thâm nhiễm nhu mô phổi hai bên. Đám mờ đông đặc nhu mô phổi phải nghĩ đến đụng dập nhu mô phổi. Gãy cung xương sườn 2 bên trái.\\
        & (\textit{Bilateral pulmonary parenchymal infiltrates. Consolidation opacity in the right lung parenchyma, suggestive of pulmonary contusion. Fracture of the left 2nd rib arch.})\\ \hdashline
        \multicolumn{1}{:c:}{\multirow{1}{*}{\textbf{X-ray}}}
        & \multicolumn{1}{c:}{\includegraphics[width=0.12\linewidth]{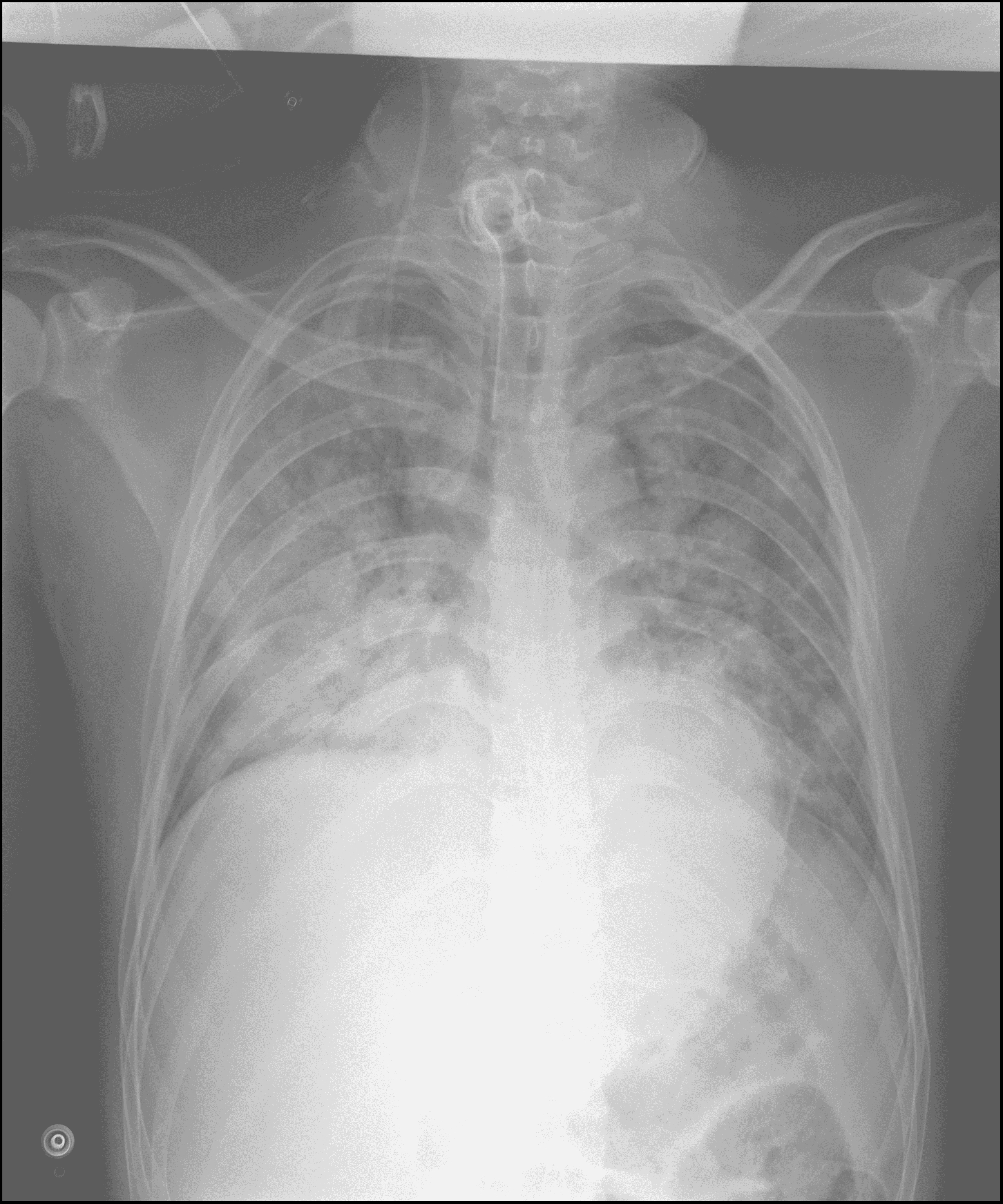}} \\
    \hdashline
    
    \end{tabular}
    
    \caption{One example from our dataset, ViX-Ray.}
    \label{tab:ViX-RayExample}
\end{table*}
In this section, we describe the data collection process, including a brief overview of the data fields in our dataset. We also analyze the anatomical parts mentioned by doctors in the findings, as well as the frequency of their diagnoses.
\subsection{Data Collection}
ViX-Ray was collected from examination records of patients at Vietnam Military Hospital 175, comprising 5,400 chest X-ray images, each accompanied by detailed findings and diagnostic impressions provided by medical specialists. To protect patient confidentiality \citep{law_of_exam_treatement}, all protected health information (PHI) \citep{isola2023protected} has been removed to ensure data privacy and security. However, clinically relevant metadata such as age and gender are retained to support diagnostic and analytical tasks. An example from the dataset is presented in Table \ref{tab:ViX-RayExample}.

\subsection{Data Analysis}
\label{subsec:Data Analysis}
In this section, we provide an in-depth analysis of the key characteristics of the ViX-Ray dataset, covering the most frequently examined anatomical regions and the diagnostic conclusions provided by medical specialists based on X-ray images. This analysis aims to offer the research community a comprehensive overview of the structure and clinical relevance of the dataset.

\begin{figure*}[t]
\begin{subfigure}{.5\textwidth}
  \centering
  \includegraphics[width=0.85\linewidth]{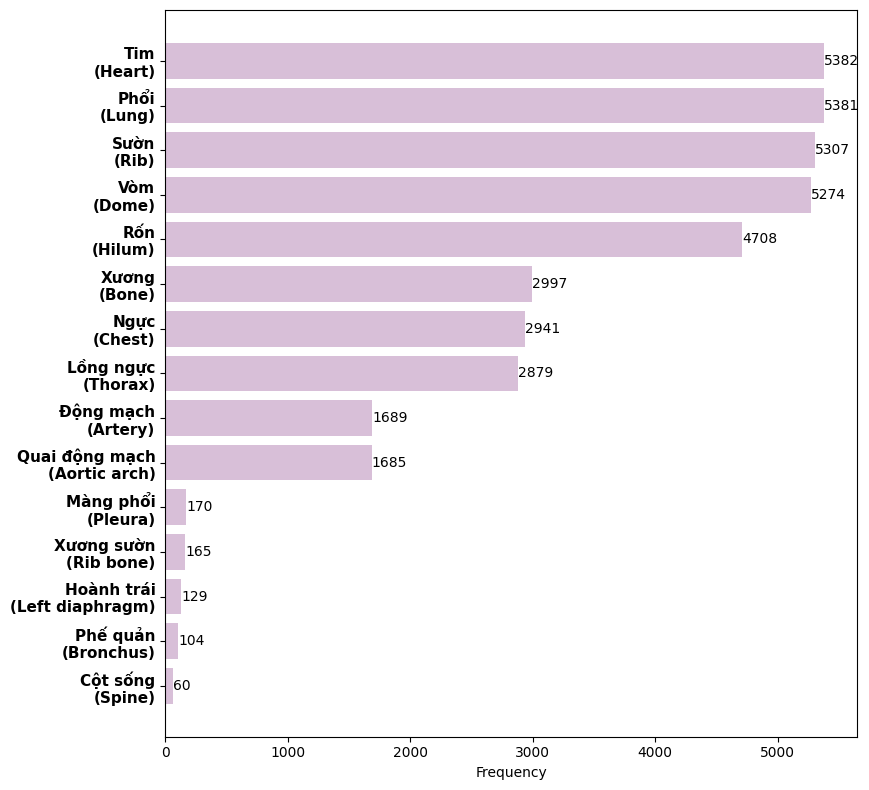}
  \caption{Frequency of anatomical parts analyzed by doctors.}
  \label{fig:BodyPartsFrequency}
\end{subfigure}%
\begin{subfigure}{.5\textwidth}
  \centering
  \includegraphics[width=1.03\linewidth]{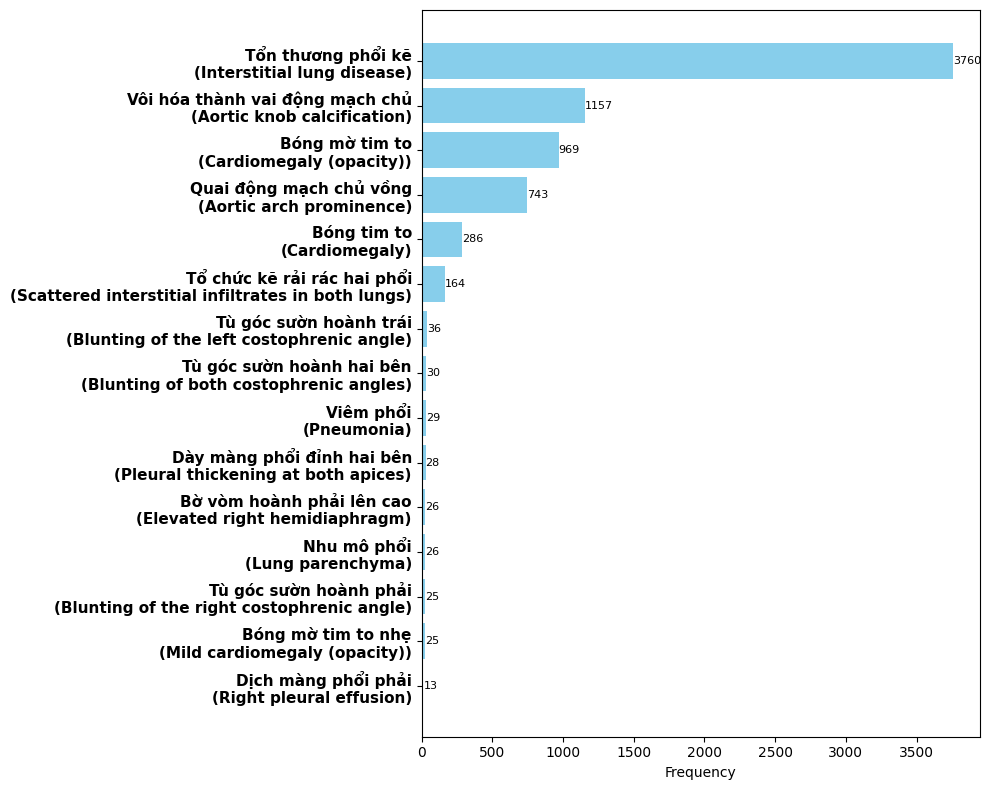}
  \caption{Frequency of diagnoses made by doctors.}
  \label{fig:DiagnosesFrequency}
\end{subfigure}
\caption{Visualization of clinical features present in the ViX-Ray dataset.}
\end{figure*}

\textbf{Body Parts Frequency:}
The medical findings in the dataset are written as descriptive narratives regarding the condition of the patient (see Table \ref{tab:ViX-RayExample} for more details). To analyze them, we utilize Stanza \citep{qi2020stanza} to generate syntactic parse trees, allowing us to extract noun phrases from the findings and count their frequency. We then filter for nouns or noun phrases most relevant to anatomical body parts and visualize the results in Figure \ref{fig:BodyPartsFrequency}. As shown in the figure, the heart (\textit{tim}) and lungs (\textit{phổi}) are the two most frequently mentioned organs in physician assessments, followed by structures such as the ribs, diaphragm dome, and pulmonary hilum. Notably, the medical reports often include not only the presence of abnormalities but also their specific locations and conditions, such as "xương sườn 2 bên trái" (left second rib arch), "Gãy cung sau xương sườn III" (posterior fracture of the third rib), or "cạnh rốn phổi trái" (near the left hilar region). This level of detail significantly increases the complexity of the dataset, as it requires models to accurately detect anatomical entities along with fine-grained positional and descriptive attributes, thus demanding a deeper understanding of human anatomical structure.

\textbf{Diagnosis Frequency:}
Similar to the previously described information extraction steps, we applied frequency analysis on the diagnoses provided by doctors, and the results are illustrated in Figure \ref{fig:DiagnosesFrequency}. From the figure, it can be observed that diagnoses related to the lungs and heart appear with high frequency—typical examples include "tổn thương phổi kẽ" (Interstitial lung disease) and "Bóng tim to" (Cardiomegaly). In addition, the specialists also provided severity levels of the conditions observed in patients. For instance, in the case of "Bóng tim to" (Cardiomegaly), a milder form is also noted as "Bóng mờ tim to nhẹ" (Mild cardiomegaly (opacity)). This presents a significant challenge for models, as they must not only accurately identify the location and characteristics of anatomical structures but also detect and classify the presence and severity of abnormalities across patients of different ages.

\subsection{Data Statistics}
\label{subsec:Data Statistic}
The ViX-Ray dataset consists of 5,400 samples, divided into training, development (dev), and test sets in an 8:1:1 ratio, as detailed in Table \ref{tab:ViX-ray Data Statistic}. We compute the minimum, maximum, and average lengths of impressions and findings after segmentation using VnCoreNLP \citep{vncorenlp}, along with the average patient age in each subset. The results show consistent distributions of linguistic and demographic features, supporting balanced and reliable experimental evaluations.
\begin{table}[h]
\centering
\resizebox{\linewidth}{!}{
\begin{tabular}{|cl|c|c|c|}
\hline
\multicolumn{2}{|l|}{} & \textbf{Train} & \textbf{Development} & \textbf{Test} \\ \hline
\multicolumn{2}{|c|}{\textbf{Num. Sample}} & 4320 & 520 & 520 \\ \hline
\multicolumn{2}{|c|}{\textbf{Avg. Age}} & 69 & 70 & 70 \\ \hline
\multicolumn{1}{|c|}{\multirow{3}{*}{\textbf{\shortstack{Findings\\Length}}}} & Min & 19 & 26 & 26 \\  
\multicolumn{1}{|c|}{} & Avg. & 46 & 45 & 46 \\  
\multicolumn{1}{|c|}{} & Max & 104 & 91 & 85 \\ \hline
\multicolumn{1}{|c|}{\multirow{3}{*}{\textbf{\shortstack{Impressions\\Length}}}} & Min & 4 & 5 & 5 \\ 
\multicolumn{1}{|c|}{} & Avg. & 12 & 14 & 13 \\ 
\multicolumn{1}{|c|}{} & Max & 67 & 52 & 45 \\ \hline
\end{tabular}
}
\caption{ViX-Ray Data Statistic}
\label{tab:ViX-ray Data Statistic}
\end{table}

\begin{table*}[t]
\centering
\resizebox{\textwidth}{!}{%

\begin{tabular}{|cl|cccccc|cccccc|}
\hline
\multicolumn{2}{|l|}{\multirow{3}{*}{\textbf{}}} & \multicolumn{6}{c|}{\textbf{Dev}} & \multicolumn{6}{c|}{\textbf{Test}} \\ \cline{3-14} 
\multicolumn{2}{|l|}{} & \multicolumn{3}{c|}{\textbf{Rouge}} & \multicolumn{1}{c|}{\multirow{2}{*}{\textbf{Bleu}}} & \multicolumn{1}{c|}{\multirow{2}{*}{\textbf{Precision}}} & \multicolumn{1}{c|}{\multirow{2}{*}{\textbf{Recall}}} & \multicolumn{3}{c|}{\textbf{Rouge}} & \multicolumn{1}{c|}{\multirow{2}{*}{\textbf{Bleu}}} & \multicolumn{1}{c|}{\multirow{2}{*}{\textbf{Precision}}} & \multicolumn{1}{c|}{\multirow{2}{*}{\textbf{Recall}}} \\ \cline{3-5} \cline{9-11}

\multicolumn{2}{|l|}{} & \multicolumn{1}{c|}{\textbf{1}} & \multicolumn{1}{c|}{\textbf{2}} & \multicolumn{1}{c|}{\textbf{L}} & \multicolumn{1}{c|}{} & \multicolumn{1}{c|}{} & \multicolumn{1}{c|}{} & \multicolumn{1}{c|}{\textbf{1}} & \multicolumn{1}{c|}{\textbf{2}} & \multicolumn{1}{c|}{\textbf{L}} & \multicolumn{1}{c|}{} & \multicolumn{1}{c|}{} & \multicolumn{1}{c|}{} \\ \hline

\multicolumn{14}{|l|}{\textbf{Stage 1 - Findings Generation}}
\\
\hline

 \multicolumn{1}{|c|}{\multirow{2}{*}{\textbf{Monolingual}}} & Vintern-v3.5 & \multicolumn{1}{c|}{81.79} & \multicolumn{1}{c|}{71.71} & \multicolumn{1}{c|}{75.99} & \multicolumn{1}{c|}{61.91} & \multicolumn{1}{c|}{57.25} & \multicolumn{1}{c|}{56.74}  & \multicolumn{1}{c|}{81.90} & \multicolumn{1}{c|}{72.10} & \multicolumn{1}{c|}{76.21} & \multicolumn{1}{c|}{61.39} & \multicolumn{1}{c|}{56.25} & \multicolumn{1}{c|}{56.12}  \\ 

 \multicolumn{1}{|c|}{} & LaVy & \multicolumn{1}{c|}{81.95} & \multicolumn{1}{c|}{71.08} & \multicolumn{1}{c|}{76.91} & \multicolumn{1}{c|}{69.11} & \multicolumn{1}{c|}{61.54} & \multicolumn{1}{c|}{61.12} & \multicolumn{1}{c|}{82.12} & \multicolumn{1}{c|}{71.35} & \multicolumn{1}{c|}{77.25} & \multicolumn{1}{c|}{69.58} & \multicolumn{1}{c|}{62.48} & \multicolumn{1}{c|}{61.52}  \\ \cline{1-14} 

 \multicolumn{1}{|c|}{\multirow{4}{*}{\textbf{Multilingual}}} & InternVL2.5 - 1B & \multicolumn{1}{c|}{66.29} & \multicolumn{1}{c|}{47.18} & \multicolumn{1}{c|}{51.06} & \multicolumn{1}{c|}{62.09} & \multicolumn{1}{c|}{56.24} & \multicolumn{1}{c|}{55.12} & \multicolumn{1}{c|}{65.67} & \multicolumn{1}{c|}{46.82} & \multicolumn{1}{c|}{50.57} & \multicolumn{1}{c|}{61.94} & \multicolumn{1}{c|}{56.24} & \multicolumn{1}{c|}{55.12} \\ 
 \multicolumn{1}{|c|}{} & Qwen2.5VL-2B & \multicolumn{1}{c|}{83.40} & \multicolumn{1}{c|}{75.25} & \multicolumn{1}{c|}{79.03} & \multicolumn{1}{c|}{69.42} & \multicolumn{1}{c|}{61.42} & \multicolumn{1}{c|}{60.24}  & \multicolumn{1}{c|}{83.87} & \multicolumn{1}{c|}{75.56} & \multicolumn{1}{c|}{79.27} & \multicolumn{1}{c|}{70.21} & \multicolumn{1}{c|}{62.21} & \multicolumn{1}{c|}{61.45} \\ 
 \multicolumn{1}{|c|}{} & Qwen2.5VL-7B & \multicolumn{1}{c|}{\textbf{84.09}} & \multicolumn{1}{c|}{\textbf{76.77}} & \multicolumn{1}{c|}{\textbf{81.11}} & \multicolumn{1}{c|}{\textbf{70.11}} & \multicolumn{1}{c|}{\textbf{68.91}} & \multicolumn{1}{c|}{\textbf{69.94}} & \multicolumn{1}{c|}{\textbf{84.30}} & \multicolumn{1}{c|}{\textbf{76.10}} & \multicolumn{1}{c|}{\textbf{81.21}} & \multicolumn{1}{c|}{\textbf{71.22}} & \multicolumn{1}{c|}{\textbf{70.51}} & \multicolumn{1}{c|}{\textbf{70.21}} \\ 
 \multicolumn{1}{|c|}{} & MiniCPM-V & \multicolumn{1}{c|}{83.21} & \multicolumn{1}{c|}{73.21} & \multicolumn{1}{c|}{79.98} & \multicolumn{1}{c|}{69.24} & \multicolumn{1}{c|}{67.56} & \multicolumn{1}{c|}{66.12}  & \multicolumn{1}{c|}{83.25} & \multicolumn{1}{c|}{72.97} & \multicolumn{1}{c|}{79.81} & \multicolumn{1}{c|}{70.12} & \multicolumn{1}{c|}{68.12} & \multicolumn{1}{c|}{65.25}  \\ \hline 
\multicolumn{14}{|l|}{\textbf{Stage 2 - Impressions Generation}}
\\ \hline

\multicolumn{1}{|c|}{\multirow{2}{*}{\textbf{Monolingual}}} & Vintern-v3.5 & \multicolumn{1}{c|}{62.82} & \multicolumn{1}{c|}{52.16} & \multicolumn{1}{c|}{57.98} & \multicolumn{1}{c|}{51.13} & \multicolumn{1}{c|}{56.74} & \multicolumn{1}{c|}{49.06} & \multicolumn{1}{c|}{60.83} & \multicolumn{1}{c|}{50.85} & \multicolumn{1}{c|}{56.25} & \multicolumn{1}{c|}{51.91} & \multicolumn{1}{c|}{56.41} & \multicolumn{1}{c|}{50.14}  \\ 

\multicolumn{1}{|c|}{} & LaVy & \multicolumn{1}{c|}{70.14} & \multicolumn{1}{c|}{62.15} & \multicolumn{1}{c|}{68.45} & \multicolumn{1}{c|}{57.02} & \multicolumn{1}{c|}{57.49} & \multicolumn{1}{c|}{51.21}  & \multicolumn{1}{c|}{70.25} & \multicolumn{1}{c|}{61.58} & \multicolumn{1}{c|}{67.78} & \multicolumn{1}{c|}{57.45} & \multicolumn{1}{c|}{57.65} & \multicolumn{1}{c|}{52.74} \\ \cline{1-14}

\multicolumn{1}{|c|}{\multirow{4}{*}{\textbf{Multilingual}}} & InternVL2.5-1B & \multicolumn{1}{c|}{61.65} & \multicolumn{1}{c|}{41.08} & \multicolumn{1}{c|}{54.22} & \multicolumn{1}{c|}{52.10} & \multicolumn{1}{c|}{51.21} & \multicolumn{1}{c|}{48.51} & \multicolumn{1}{c|}{69.05} & \multicolumn{1}{c|}{42.79} & \multicolumn{1}{c|}{56.22} & \multicolumn{1}{c|}{\textbf{62.02}} & \multicolumn{1}{c|}{52.35} & \multicolumn{1}{c|}{49.32} \\ 

\multicolumn{1}{|c|}{} & Qwen2.5VL-2B & \multicolumn{1}{c|}{73.26} & \multicolumn{1}{c|}{64.31} & \multicolumn{1}{c|}{70.81} & \multicolumn{1}{c|}{57.11} & \multicolumn{1}{c|}{54.38} & \multicolumn{1}{c|}{53.73}  & \multicolumn{1}{c|}{71.75} & \multicolumn{1}{c|}{62.48} & \multicolumn{1}{c|}{69.27} & \multicolumn{1}{c|}{57.39} & \multicolumn{1}{c|}{55.01} & \multicolumn{1}{c|}{53.12}  \\ 

\multicolumn{1}{|c|}{} & Qwen2.5VL-7B & \multicolumn{1}{c|}{\textbf{74.17}} & \multicolumn{1}{c|}{\textbf{65.75}} & \multicolumn{1}{c|}{\textbf{71.81}} & \multicolumn{1}{c|}{\textbf{60.11}} & \multicolumn{1}{c|}{\textbf{60.58}} & \multicolumn{1}{c|}{\textbf{61.94}}  & \multicolumn{1}{c|}{\textbf{73.89}} & \multicolumn{1}{c|}{\textbf{64.66}} & \multicolumn{1}{c|}{\textbf{71.11}} & \multicolumn{1}{c|}{59.56} & \multicolumn{1}{c|}{\textbf{61.25}} & \multicolumn{1}{c|}{\textbf{62.14}} \\ 

\multicolumn{1}{|c|}{} & MiniCPM-V & \multicolumn{1}{c|}{71.12} & \multicolumn{1}{c|}{63.31} & \multicolumn{1}{c|}{69.28} & \multicolumn{1}{c|}{58.12} & \multicolumn{1}{c|}{58.49} & \multicolumn{1}{c|}{50.84} & \multicolumn{1}{c|}{70.16} & \multicolumn{1}{c|}{61.97} & \multicolumn{1}{c|}{68.25} & \multicolumn{1}{c|}{57.25} & \multicolumn{1}{c|}{57.48} & \multicolumn{1}{c|}{40.22}  \\ \hline
\end{tabular}%
}
\caption{Fine-tuning results of VLMs in \textbf{Stage 1 – findings generation} and \textbf{Stage 2 – impressions generation} (\%). We present only the results obtained after fine-tuning. }
\label{tab:stage1_and_stage2_only_fine_tune}
\end{table*}

\section{Experiments and Results}
\subsection{Baseline Models}
For the baseline models, we utilize both multilingual and monolingual Vision-Language Models (VLMs) with various model sizes. However, due to resource constraints, we only use versions with fewer than 7 billion parameters—for example, for Qwen2.5-VL \citep{qwen2.5-vl}, we limit our experiments to the 2B and 7B instruct versions.

\textbf{Monolingual Vision Language Models:} We employ two Vision-Language Models (VLMs), namely Lavy \citep{lavy} and Vintern \citep{vintern}. Lavy, introduced by \citeauthor{lavy}, is built on a hybrid architecture combining a CLIP-Large vision encoder \citep{CLIP} with a Vietnamese monolingual language model, Vistral-7B \citep{vistral}. The two modalities are integrated using two MLP layers that project visual features into the embedding space of the language model. Vintern \citep{vintern}, on the other hand, utilizes InternViT \citep{internvl-1.5} as the vision encoder to extract visual features, and a multilingual LLM — Qwen2-0.5B-Instruct \citep{qwen2technicalreport} — as the language decoder. Similar to Lavy, it also uses two MLP projection layers to align the vision and language representations. Both models are trained on large-scale Vietnamese data. For Lavy, the training corpus includes English-translated datasets such as LAION-CC-SBU \citep{visual_instruction_tuning} and GPT-generated multimodal instructions. Vintern, by contrast, is trained on 15 diverse Vietnamese datasets covering a range of tasks from general visual QA \citep{vista, openvivqa}, document QA \citep{vintern}, to handwriting QA \citep{Uit-hwdb}. This extensive training enables both models to effectively handle vision-language tasks in the Vietnamese language, particularly those involving visual question answering (VQA).

\textbf{Multilingual Vision Language Models}: For multilingual VLMs, we utilize three model architectures: InternVL 2.5, MiniCPM-V 2.6, and Qwen2.5-VL \citep{qwen2.5-vl}. InternVL 2.5 \citep{internvl-2.5} is an enhanced version of its predecessor 2.0, maintaining the 'Vision-MLP-LLM' architecture widely adopted in previous research \citep{liu2024improved, chen2024internvl, zhu2024minigpt, lu2024deepseek}. It incorporates an incremental training strategy similar to that applied in version 1.5, including dynamic resolution training, which enhances the ability of the model to extract open-ended features and adapt to real-world scenarios. Qwen2.5-VL, a multilingual VLM introduced by \citeauthor{qwen2.5-vl}, can process images at native resolutions and handle varying video frame rates. This is achieved through window attention across most layers, combined with RoPE and its multimodal extension, MRoPE, which enhances temporal understanding and increases robustness in real-world applications. MiniCPM-V 2.6 \citep{minicpm-v}, another multilingual VLM with Vietnamese language support, follows a lightweight design philosophy aimed at on-device deployment. It can handle high-resolution images (e.g., 1344×1344 pixels) and exhibits reduced hallucination rates by incorporating RLAIF-V \citep{RLAIF-V} and RLHF-V \citep{rlhf-v}.

\subsection{Evaluation Metrics}
\label{subsec:Evaluation_Metrics}
To evaluate the performance of the VLM, we employ two main groups of metrics: lexical metrics, which assess the fluency and domain alignment of the generated text in the medical context, and precision-based metrics, which assess the factual accuracy of the generated information.

\textbf{Lexical Evaluation:} For lexical metrics, we use two standard measures: ROUGE and BLEU. Specifically, we adopt ROUGE-1, ROUGE-2, and ROUGE-L to evaluate the overlap of unigrams, bigrams, and the longest common subsequence (LCS) between the generated text and the reference annotations. These metrics help capture surface-level similarity and fluency within the generated responses.

\textbf{Precision Evaluation:} To assess the factual accuracy of the generated content, we draw inspiration from prior work on decomposing factual information from claims verification tasks \citep{min2023factscore, wang2025openfactcheck, li2025loki}, where the faithfulness of information is evaluated against a trustworthy context. Based on these concepts, we utilize a large language model (GPT-4o) to decompose both the generated text (denoted as $\mathcal{T}$) and the ground truth (denoted as $\mathcal{G}$) into sets of $n$ atomic facts. Let $\mathcal{T} = \{t_1, t_2, \ldots, t_n\}$ be the set of atomic facts from the generated text, and $\mathcal{G} = \{g_1, g_2, \ldots, g_m\}$ be the set of atomic facts from the ground truth. We then compute the factual precision as the ratio of atomic facts in $\mathcal{T}$ that also appear in $\mathcal{G}$:

\[
\text{Precision} = \frac{|\mathcal{T} \cap \mathcal{G}|}{|\mathcal{T}|}
\]

In this equation, $|\mathcal{T} \cap \mathcal{G}|$ denotes the number of atomic facts correctly matched between the generated text and the ground truth, and $|\mathcal{T}|$ is the total number of atomic facts in the generated text.

\textbf{Recall Evaluation:}  Analogous to our precision evaluation, we assess recall to determine the completeness of the factual information captured by the model. We decompose both the generated text ($\mathcal{T}$) and the ground truth ($\mathcal{G}$) into sets of atomic facts. Recall is then calculated as the proportion of ground-truth atomic facts that are correctly represented in the generated text:
\[
\text{Recall} = \frac{|\mathcal{T} \cap \mathcal{G}|}{|\mathcal{G}|}
\]
This metric highlights the extent to which the model captures all relevant information, offering insight into any omissions in the generated content.

\subsection{Experiment Setup}
\label{subsec:Experiment_Setup}
Since the dataset includes two distinct textual components, three experiments are conducted. Stage 1 and Stage 2 focus on visual instruction tuning of vision-language models (VLMs) for generating findings and impressions from chest X-ray images. The final experiment, Stage 3, involves multi-turn visual instruction tuning, where the model first generates findings and then derives impressions based on them.

\textbf{Stage 1 – Findings Generation:} In this stage, VLMs are visually fine-tuned to generate findings. A prompt is constructed using patient metadata such as age, gender, and view type. An example of the prompt format is shown below:

\begin{table}[h]
\centering
\begin{tabular}{p{0.9\linewidth}}
\textbf{</text>} Ảnh chụp X-ray <\textit{\textbf{View}}> (<\textit{\textbf{View Definition}}>) bệnh nhân <\textit{\textbf{Gender}}>, <\textit{\textbf{Age}}> tuổi. Cho biết bệnh nhân bị gì? \textbf{</text>}
\end{tabular}
\label{tab:my_label}
\end{table}
Each prompt is appended with the corresponding X-ray image. A full example is provided in Appendix \ref{sec:appendixA1}.

\textbf{Stage 2 – Impressions Generation:} Following the same methodology as Stage 1, VLMs are fine-tuned to generate impressions. The prompt format remains consistent, and a detailed example is included in Appendix \ref{sec:appendixA2}.

\textbf{Stage 3 – Multi-turn Generation:} This experiment is inspired by the typical diagnostic process of clinicians, who begin by reviewing the condition of the patient before providing a final clinical impression. First, the model is prompted to generate findings based on the chest X-ray, using the same prompt structure as in the previous stages. Then, using the generated findings, the model is asked to produce an impression. This multi-turn setup encourages the model to perform a more comprehensive analysis of the X-ray image and formulate deeper, more informed clinical conclusions. A full example of the multi-turn prompt is available in Appendix \ref{sec:appendixA3}.

\begin{table*}[t]
\centering
\resizebox{\textwidth}{!}{
\begin{tabular}{|clc|cccccc|cccccc|}
\hline
\multicolumn{3}{|l|}{\multirow{3}{*}{}} & \multicolumn{6}{c|}{\textbf{Dev}} & \multicolumn{6}{c|}{\textbf{Test}} \\ \cline{4-15}
\multicolumn{3}{|l|}{} & \multicolumn{3}{c|}{\textbf{Rouge}} & \multicolumn{1}{c|}{\multirow{2}{*}{\textbf{Bleu}}} & \multicolumn{1}{c|}{\multirow{2}{*}{\textbf{Precision}}} & \multirow{2}{*}{\textbf{Recall}} & \multicolumn{3}{c|}{\textbf{Rouge}} & \multicolumn{1}{c|}{\multirow{2}{*}{\textbf{Bleu}}} & \multicolumn{1}{c|}{\multirow{2}{*}{\textbf{Precision}}} & \multirow{2}{*}{\textbf{Recall}} \\ \cline{4-6} \cline{10-12}
\multicolumn{3}{|l|}{} & \textbf{1} & \textbf{2} & \multicolumn{1}{c|}{\textbf{L}} & \multicolumn{1}{c|}{} & \multicolumn{1}{c|}{} &  & \textbf{1} & \textbf{2} & \multicolumn{1}{c|}{\textbf{L}} & \multicolumn{1}{c|}{} & \multicolumn{1}{c|}{} &  \\ \hline
\multicolumn{1}{|c|}{\multirow{8}{*}{\textbf{Monolingual}}} & \multicolumn{1}{l|}{\multirow{4}{*}{Vintern-v3.5}} & \multirow{2}{*}{Findings} & \multicolumn{1}{c|}{42.84\%} & \multicolumn{1}{c|}{25.14\%} & \multicolumn{1}{c|}{31.98\%} & \multicolumn{1}{c|}{48.53\%} & \multicolumn{1}{c|}{47.54\%} & 46.48\% & \multicolumn{1}{c|}{43.87\%} & \multicolumn{1}{c|}{26.38\%} & \multicolumn{1}{c|}{33.07\%} & \multicolumn{1}{c|}{48.23\%} & \multicolumn{1}{c|}{46.52\%} & 45.12\% \\ 
\multicolumn{1}{|c|}{} & \multicolumn{1}{l|}{} &  & \multicolumn{1}{c|}{\textcolor{red}{\textdownarrow38.95\%}} & \multicolumn{1}{c|}{\textcolor{red}{\textdownarrow46.57\%}} & \multicolumn{1}{c|}{\textcolor{red}{\textdownarrow44.01\%}} & \multicolumn{1}{c|}{\textcolor{red}{\textdownarrow13.38\%}} & \multicolumn{1}{c|}{\textcolor{red}{\textdownarrow9.71\%}} & \textcolor{red}{\textdownarrow 10.26\%} & \multicolumn{1}{c|}{\textcolor{red}{\textdownarrow38.03\%}} & \multicolumn{1}{c|}{\textcolor{red}{\textdownarrow45.72\%}} & \multicolumn{1}{c|}{\textcolor{red}{\textdownarrow43.14\%}} & \multicolumn{1}{c|}{\textcolor{red}{\textdownarrow13.16\%}} & \multicolumn{1}{c|}{\textcolor{red}{\textdownarrow9.73\%}} & \textcolor{red}{\textdownarrow 11.00\%} \\ \cdashline{3-15}
\multicolumn{1}{|c|}{} & \multicolumn{1}{l|}{} & \multirow{2}{*}{Impression} & \multicolumn{1}{c|}{49.36\%} & \multicolumn{1}{c|}{29.23\%} & \multicolumn{1}{c|}{43.86\%} & \multicolumn{1}{c|}{42.36\%} & \multicolumn{1}{c|}{47.51\%} & 48.12\% & \multicolumn{1}{c|}{49.93\%} & \multicolumn{1}{c|}{29.84\%} & \multicolumn{1}{c|}{44.29\%} & \multicolumn{1}{c|}{43.84\%} & \multicolumn{1}{c|}{46.85\%} & 47.25\% \\ 
\multicolumn{1}{|c|}{} & \multicolumn{1}{l|}{} &  & \multicolumn{1}{c|}{\textcolor{red}{\textdownarrow13.46\%}} & \multicolumn{1}{c|}{\textcolor{red}{\textdownarrow22.93\%}} & \multicolumn{1}{c|}{\textcolor{red}{\textdownarrow14.12\%}} & \multicolumn{1}{c|}{\textcolor{red}{\textdownarrow8.77\%}} & \multicolumn{1}{c|}{\textcolor{red}{\textdownarrow9.23\%}} & \textcolor{red}{\textdownarrow 0.94\%} & \multicolumn{1}{c|}{\textcolor{red}{\textdownarrow10.90\%}} & \multicolumn{1}{c|}{\textcolor{red}{\textdownarrow21.01\%}} & \multicolumn{1}{c|}{\textcolor{red}{\textdownarrow11.96\%}} & \multicolumn{1}{c|}{\textcolor{red}{\textdownarrow8.07\%}} & \multicolumn{1}{c|}{\textcolor{red}{\textdownarrow9.56\%}} & \textcolor{red}{\textdownarrow 2.89\%} \\ \cline{2-15} 
\multicolumn{1}{|c|}{} & \multicolumn{1}{l|}{\multirow{4}{*}{LaVy}} & \multirow{2}{*}{Findings} & \multicolumn{1}{c|}{82.96\%} & \multicolumn{1}{c|}{74.02\%} & \multicolumn{1}{c|}{76.21\%} & \multicolumn{1}{c|}{66.92\%} & \multicolumn{1}{c|}{55.41\%} & 55.12\% & \multicolumn{1}{c|}{83.51\%} & \multicolumn{1}{c|}{73.89\%} & \multicolumn{1}{c|}{75.85\%} & \multicolumn{1}{c|}{67.54\%} & \multicolumn{1}{c|}{54.98\%} & 54.52\% \\ 
\multicolumn{1}{|c|}{} & \multicolumn{1}{l|}{} &  & \multicolumn{1}{c|}{\textcolor{blue}{\textuparrow 1.01\%}} & \multicolumn{1}{c|}{\textcolor{blue}{\textuparrow 2.94\%}} & \multicolumn{1}{c|}{\textcolor{red}{\textdownarrow0.70\%}} & \multicolumn{1}{c|}{\textcolor{red}{\textdownarrow2.19\%}} & \multicolumn{1}{c|}{\textcolor{red}{\textdownarrow6.13\%}} & \textcolor{red}{\textdownarrow 6.00\%} & \multicolumn{1}{c|}{\textcolor{blue}{\textuparrow 1.39\%}} & \multicolumn{1}{c|}{\textcolor{blue}{\textuparrow 2.54\%}} & \multicolumn{1}{c|}{\textcolor{red}{\textdownarrow1.40\%}} & \multicolumn{1}{c|}{\textcolor{red}{\textdownarrow2.04\%}} & \multicolumn{1}{c|}{\textcolor{red}{\textdownarrow7.50\%}} & \textcolor{red}{\textdownarrow 7.00\%} \\ \cdashline{3-15}
\multicolumn{1}{|c|}{} & \multicolumn{1}{l|}{} & \multirow{2}{*}{Impression} & \multicolumn{1}{c|}{70.12\%} & \multicolumn{1}{c|}{59.84\%} & \multicolumn{1}{c|}{67.21\%} & \multicolumn{1}{c|}{55.07\%} & \multicolumn{1}{c|}{41.25\%} & 56.52\% & \multicolumn{1}{c|}{69.52\%} & \multicolumn{1}{c|}{60.12\%} & \multicolumn{1}{c|}{66.52\%} & \multicolumn{1}{c|}{54.74\%} & \multicolumn{1}{c|}{40.95\%} & 55.41\% \\ 
\multicolumn{1}{|c|}{} & \multicolumn{1}{l|}{} &  & \multicolumn{1}{c|}{\textcolor{red}{\textdownarrow0.02\%}} & \multicolumn{1}{c|}{\textcolor{red}{\textdownarrow2.31\%}} & \multicolumn{1}{c|}{\textcolor{red}{\textdownarrow1.24\%}} & \multicolumn{1}{c|}{\textcolor{red}{\textdownarrow1.95\%}} & \multicolumn{1}{c|}{\textcolor{red}{\textdownarrow16.24\%}} & \textcolor{blue}{\textuparrow 5.31\%} & \multicolumn{1}{c|}{\textcolor{red}{\textdownarrow0.73\%}} & \multicolumn{1}{c|}{\textcolor{red}{\textdownarrow1.46\%}} & \multicolumn{1}{c|}{\textcolor{red}{\textdownarrow1.26\%}} & \multicolumn{1}{c|}{\textcolor{red}{\textdownarrow2.71\%}} & \multicolumn{1}{c|}{\textcolor{red}{\textdownarrow16.70\%}} & \textcolor{blue}{\textuparrow 2.67\%} \\ \hline
\multicolumn{1}{|c|}{\multirow{16}{*}{\textbf{Multilingual}}} & \multicolumn{1}{l|}{\multirow{4}{*}{InternVL2.5}} & \multirow{2}{*}{Findings} & \multicolumn{1}{c|}{56.16\%} & \multicolumn{1}{c|}{33.21\%} & \multicolumn{1}{c|}{40.48\%} & \multicolumn{1}{c|}{57.61\%} & \multicolumn{1}{c|}{41.54\%} & 42.01\% & \multicolumn{1}{c|}{55.87\%} & \multicolumn{1}{c|}{33.49\%} & \multicolumn{1}{c|}{40.32\%} & \multicolumn{1}{c|}{64.01\%} & \multicolumn{1}{c|}{41.33\%} & 42.51\% \\ 
\multicolumn{1}{|c|}{} & \multicolumn{1}{l|}{} &  & \multicolumn{1}{c|}{\textcolor{red}{\textdownarrow10.13\%}} & \multicolumn{1}{c|}{\textcolor{red}{\textdownarrow13.97\%}} & \multicolumn{1}{c|}{\textcolor{red}{\textdownarrow10.58\%}} & \multicolumn{1}{c|}{\textcolor{red}{\textdownarrow4.48\%}} & \multicolumn{1}{c|}{\textcolor{red}{\textdownarrow14.70\%}} & \textcolor{red}{\textdownarrow 13.11\%} & \multicolumn{1}{c|}{\textcolor{red}{\textdownarrow9.80\%}} & \multicolumn{1}{c|}{\textcolor{red}{\textdownarrow13.33\%}} & \multicolumn{1}{c|}{\textcolor{red}{\textdownarrow10.25\%}} & \multicolumn{1}{c|}{\textcolor{blue}{\textuparrow 2.08\%}} & \multicolumn{1}{c|}{\textcolor{red}{\textdownarrow14.91\%}} & \textcolor{red}{\textdownarrow 12.61\%} \\ \cdashline{3-15}
\multicolumn{1}{|c|}{} & \multicolumn{1}{l|}{} & \multirow{2}{*}{Impression} & \multicolumn{1}{c|}{39.64\%} & \multicolumn{1}{c|}{17.41\%} & \multicolumn{1}{c|}{30.47\%} & \multicolumn{1}{c|}{44.34\%} & \multicolumn{1}{c|}{42.45\%} & 41.55\% & \multicolumn{1}{c|}{39.07\%} & \multicolumn{1}{c|}{16.38\%} & \multicolumn{1}{c|}{30.20\%} & \multicolumn{1}{c|}{54.09\%} & \multicolumn{1}{c|}{42.25\%} & 41.18\% \\ 
\multicolumn{1}{|c|}{} & \multicolumn{1}{l|}{} &  & \multicolumn{1}{c|}{\textcolor{red}{\textdownarrow22.01\%}} & \multicolumn{1}{c|}{\textcolor{red}{\textdownarrow23.67\%}} & \multicolumn{1}{c|}{\textcolor{red}{\textdownarrow23.75\%}} & \multicolumn{1}{c|}{\textcolor{red}{\textdownarrow17.76\%}} & \multicolumn{1}{c|}{\textcolor{red}{\textdownarrow8.76\%}} & \textcolor{red}{\textdownarrow 6.96\%} & \multicolumn{1}{c|}{\textcolor{red}{\textdownarrow29.98\%}} & \multicolumn{1}{c|}{\textcolor{red}{\textdownarrow26.41\%}} & \multicolumn{1}{c|}{\textcolor{red}{\textdownarrow26.02\%}} & \multicolumn{1}{c|}{\textcolor{red}{\textdownarrow7.93\%}} & \multicolumn{1}{c|}{\textcolor{red}{\textdownarrow10.10\%}} & \textcolor{red}{\textdownarrow 8.14\%} \\ \cline{2-15} 
\multicolumn{1}{|c|}{} & \multicolumn{1}{l|}{\multirow{4}{*}{Qwen2.5VL-2B}} & \multirow{2}{*}{Findings} & \multicolumn{1}{c|}{80.24\%} & \multicolumn{1}{c|}{71.44\%} & \multicolumn{1}{c|}{75.60\%} & \multicolumn{1}{c|}{66.06\%} & \multicolumn{1}{c|}{64.84\%} & 63.55\% & \multicolumn{1}{c|}{80.28\%} & \multicolumn{1}{c|}{71.90\%} & \multicolumn{1}{c|}{75.93\%} & \multicolumn{1}{c|}{66.86\%} & \multicolumn{1}{c|}{65.25\%} & 64.15\% \\ 
\multicolumn{1}{|c|}{} & \multicolumn{1}{l|}{} &  & \multicolumn{1}{c|}{\textcolor{red}{\textdownarrow3.16\%}} & \multicolumn{1}{c|}{\textcolor{red}{\textdownarrow3.81\%}} & \multicolumn{1}{c|}{\textcolor{red}{\textdownarrow3.43\%}} & \multicolumn{1}{c|}{\textcolor{red}{\textdownarrow3.36\%}} & \multicolumn{1}{c|}{\textcolor{blue}{\textuparrow 3.42\%}} & \textcolor{blue}{\textuparrow 3.31\%} & \multicolumn{1}{c|}{\textcolor{red}{\textdownarrow3.59\%}} & \multicolumn{1}{c|}{\textcolor{red}{\textdownarrow3.66\%}} & \multicolumn{1}{c|}{\textcolor{red}{\textdownarrow3.34\%}} & \multicolumn{1}{c|}{\textcolor{red}{\textdownarrow3.35\%}} & \multicolumn{1}{c|}{\textcolor{blue}{\textuparrow 3.04\%}} & \textcolor{blue}{\textuparrow 2.70\%} \\ \cdashline{3-15}
\multicolumn{1}{|c|}{} & \multicolumn{1}{l|}{} & \multirow{2}{*}{Impression} & \multicolumn{1}{c|}{78.07\%} & \multicolumn{1}{c|}{75.10\%} & \multicolumn{1}{c|}{76.91\%} & \multicolumn{1}{c|}{66.58\%} & \multicolumn{1}{c|}{73.23\%} & 68.71\% & \multicolumn{1}{c|}{76.13\%} & \multicolumn{1}{c|}{72.44\%} & \multicolumn{1}{c|}{75.13\%} & \multicolumn{1}{c|}{65.32\%} & \multicolumn{1}{c|}{68.34\%} & 66.20\% \\ 
\multicolumn{1}{|c|}{} & \multicolumn{1}{l|}{} &  & \multicolumn{1}{c|}{\textcolor{blue}{\textuparrow 4.81\%}} & \multicolumn{1}{c|}{\textcolor{blue}{\textuparrow 10.79\%}} & \multicolumn{1}{c|}{\textcolor{blue}{\textuparrow 6.10\%}} & \multicolumn{1}{c|}{\textcolor{blue}{\textuparrow 9.47\%}} & \multicolumn{1}{c|}{\textcolor{blue}{\textuparrow 18.85\%}} & \textcolor{blue}{\textuparrow 14.98\%} & \multicolumn{1}{c|}{\textcolor{blue}{\textuparrow 4.38\%}} & \multicolumn{1}{c|}{\textcolor{blue}{\textuparrow 9.96\%}} & \multicolumn{1}{c|}{\textcolor{blue}{\textuparrow 5.86\%}} & \multicolumn{1}{c|}{\textcolor{blue}{\textuparrow 7.93\%}} & \multicolumn{1}{c|}{\textcolor{blue}{\textuparrow 13.33\%}} & \textcolor{blue}{\textuparrow 13.08\%} \\ \cline{2-15} 

\multicolumn{1}{|c|}{} & \multicolumn{1}{l|}{\multirow{4}{*}{Qwen2.5VL-7B}} & \multirow{2}{*}{Findings} & \multicolumn{1}{c|}{\textbf{84.81\%}} & \multicolumn{1}{c|}{\textbf{76.97\%}} & \multicolumn{1}{c|}{\textbf{80.47\%}} & \multicolumn{1}{c|}{\textbf{71.35\%}} & \multicolumn{1}{c|}{\textbf{70.34\%}} & \textbf{72.28\%} & \multicolumn{1}{c|}{\textbf{84.40\%}} & \multicolumn{1}{c|}{\textbf{75.90\%}} & \multicolumn{1}{c|}{\textbf{79.60\%}} & \multicolumn{1}{c|}{\textbf{69.85\%}} & \multicolumn{1}{c|}{\textbf{69.78\%}} & \textbf{70.60\%} \\ 
\multicolumn{1}{|c|}{} & \multicolumn{1}{l|}{} &  & \multicolumn{1}{c|}{\textcolor{blue}{\textuparrow 0.72\%}} & \multicolumn{1}{c|}{\textcolor{blue}{\textuparrow 0.20\%}} & \multicolumn{1}{c|}{\textcolor{red}{\textdownarrow0.64\%}} & \multicolumn{1}{c|}{\textcolor{blue}{\textuparrow 1.24\%}} & \multicolumn{1}{c|}{\textcolor{blue}{\textuparrow 1.43\%}} & \textcolor{blue}{\textuparrow 2.34\%} & \multicolumn{1}{c|}{\textcolor{blue}{\textuparrow 0.10\%}} & \multicolumn{1}{c|}{\textcolor{red}{\textdownarrow0.20\%}} & \multicolumn{1}{c|}{\textcolor{red}{\textdownarrow1.61\%}} & \multicolumn{1}{c|}{\textcolor{red}{\textdownarrow1.37\%}} & \multicolumn{1}{c|}{\textcolor{red}{\textdownarrow0.73\%}} & \textcolor{blue}{\textuparrow 0.39\%} \\ \cdashline{3-15}
\multicolumn{1}{|c|}{} & \multicolumn{1}{l|}{} & \multirow{2}{*}{Impression} & \multicolumn{1}{c|}{\textbf{95.93\%}} & \multicolumn{1}{c|}{\textbf{94.34\%}} & \multicolumn{1}{c|}{\textbf{95.08\%}} & \multicolumn{1}{c|}{\textbf{92.04\%}} & \multicolumn{1}{c|}{\textbf{92.14\%}} & \textbf{92.11\%} & \multicolumn{1}{c|}{\textbf{95.20\%}} & \multicolumn{1}{c|}{\textbf{93.80\%}} & \multicolumn{1}{c|}{\textbf{94.68\%}} & \multicolumn{1}{c|}{\textbf{89.75\%}} & \multicolumn{1}{c|}{\textbf{89.95\%}} & \textbf{90.88\%} \\ 
\multicolumn{1}{|c|}{} & \multicolumn{1}{l|}{} &  & \multicolumn{1}{c|}{\textcolor{blue}{\textuparrow 21.76\%}} & \multicolumn{1}{c|}{\textcolor{blue}{\textuparrow 28.59\%}} & \multicolumn{1}{c|}{\textcolor{blue}{\textuparrow 23.27\%}} & \multicolumn{1}{c|}{\textcolor{blue}{\textuparrow 31.93\%}} & \multicolumn{1}{c|}{\textcolor{blue}{\textuparrow 31.56\%}} & \textcolor{blue}{\textuparrow 30.17\%} & \multicolumn{1}{c|}{\textcolor{blue}{\textuparrow 21.31\%}} & \multicolumn{1}{c|}{\textcolor{blue}{\textuparrow 29.14\%}} & \multicolumn{1}{c|}{\textcolor{blue}{\textuparrow 23.57\%}} & \multicolumn{1}{c|}{\textcolor{blue}{\textuparrow 30.19\%}} & \multicolumn{1}{c|}{\textcolor{blue}{\textuparrow 28.70\%}} & \textcolor{blue}{\textuparrow 28.74\%} \\ \cline{2-15} 

\multicolumn{1}{|c|}{} & \multicolumn{1}{l|}{\multirow{4}{*}{MiniCPM-V}} & \multirow{2}{*}{Findings} & \multicolumn{1}{c|}{84.25\%} & \multicolumn{1}{c|}{74.41\%} & \multicolumn{1}{c|}{77.61\%} & \multicolumn{1}{c|}{69.23\%} & \multicolumn{1}{c|}{65.57\%} & 67.88\% & \multicolumn{1}{c|}{84.39\%} & \multicolumn{1}{c|}{74.27\%} & \multicolumn{1}{c|}{77.44\%} & \multicolumn{1}{c|}{68.74\%} & \multicolumn{1}{c|}{64.27\%} & 66.58\% \\ 
\multicolumn{1}{|c|}{} & \multicolumn{1}{l|}{} &  & \multicolumn{1}{c|}{\textcolor{blue}{\textuparrow 1.04\%}} & \multicolumn{1}{c|}{\textcolor{blue}{\textuparrow 1.20\%}} & \multicolumn{1}{c|}{\textcolor{red}{\textdownarrow2.37\%}} & \multicolumn{1}{c|}{\textcolor{red}{\textdownarrow0.01\%}} & \multicolumn{1}{c|}{\textcolor{red}{\textdownarrow1.99\%}} & \textcolor{blue}{\textuparrow 1.76\%} & \multicolumn{1}{c|}{\textcolor{blue}{\textuparrow 1.14\%}} & \multicolumn{1}{c|}{\textcolor{blue}{\textuparrow 1.30\%}} & \multicolumn{1}{c|}{\textcolor{red}{\textdownarrow2.37\%}} & \multicolumn{1}{c|}{\textcolor{red}{\textdownarrow1.38\%}} & \multicolumn{1}{c|}{\textcolor{red}{\textdownarrow3.85\%}} & \textcolor{blue}{\textuparrow 1.33\%} \\ \cdashline{3-15}
\multicolumn{1}{|c|}{} & \multicolumn{1}{l|}{} & \multirow{2}{*}{Impression} & \multicolumn{1}{c|}{70.86\%} & \multicolumn{1}{c|}{61.95\%} & \multicolumn{1}{c|}{69.42\%} & \multicolumn{1}{c|}{57.91\%} & \multicolumn{1}{c|}{45.86\%} & 53.59\% & \multicolumn{1}{c|}{70.55\%} & \multicolumn{1}{c|}{61.25\%} & \multicolumn{1}{c|}{68.08\%} & \multicolumn{1}{c|}{56.11\%} & \multicolumn{1}{c|}{44.21\%} & 53.12\% \\ 
\multicolumn{1}{|c|}{} & \multicolumn{1}{l|}{} &  & \multicolumn{1}{c|}{\textcolor{red}{\textdownarrow0.26\%}} & \multicolumn{1}{c|}{\textcolor{red}{\textdownarrow1.36\%}} & \multicolumn{1}{c|}{\textcolor{blue}{\textuparrow 0.14\%}} & \multicolumn{1}{c|}{\textcolor{red}{\textdownarrow0.21\%}} & \multicolumn{1}{c|}{\textcolor{red}{\textdownarrow12.63\%}} & \textcolor{blue}{\textuparrow 2.75\%} & \multicolumn{1}{c|}{\textcolor{blue}{\textuparrow 0.39\%}} & \multicolumn{1}{c|}{\textcolor{red}{\textdownarrow0.72\%}} & \multicolumn{1}{c|}{\textcolor{red}{\textdownarrow0.17\%}} & \multicolumn{1}{c|}{\textcolor{red}{\textdownarrow1.14\%}} & \multicolumn{1}{c|}{\textcolor{red}{\textdownarrow13.27\%}} & \textcolor{blue}{\textuparrow 12.90\%} \\ \hline
\end{tabular}
}
\caption{Results of multi-turn visual fine-tuning on the ViX-Ray dataset (\%). In this setup, the VLMs sequentially generate findings followed by impressions. To illustrate the effect of multi-turn fine-tuning, we compare the findings generated in Stage 3 with those in Stage 1, and the impressions generated in Stage 3 with those in Stage 2. Performance improvements are highlighted in \textcolor{blue}{blue up arrow} (\textcolor{blue}{\textuparrow}), while decreases are marked in \textcolor{red}{red down arrow} (\textcolor{red}{\textdownarrow}).}
\label{tab:stage3-maintable}
\end{table*}

\subsection{Experimental Results}
\label{sec:Experimental Results}

In this section, we present and analyze the performance of the VLM models on both the development and test sets after fine-tuning on the ViX-Ray dataset. The results from Stage 1 and Stage 2 are shown in Table \ref{tab:stage1_and_stage2_only_fine_tune}, while the results from Stage 3 are illustrated in Table \ref{tab:stage3-maintable}. 

\textbf{Monolingual Result:} The results in Stage 1 – findings generation – show that both LaVy and Vintern-v3.5 are capable of producing observations similar to those written by radiologists, as reflected by BLEU and ROUGE scores. However, both models perform poorly in terms of precision and recall, indicating difficulties in capturing all the clinically relevant details. In Stage 2 - impressions generation - both models exhibit a noticeable drop in performance, particularly in lexical metrics, suggesting challenges in generating accurate impressions. Furthermore, low recall scores point to the presence of redundant or irrelevant information in their outputs. These results highlight the difficulty of our ViX-Ray dataset, which demands not only accuracy but also conciseness, posing a substantial challenge for current vision-language models.

\textbf{Multilingual Result:} The results in Table \ref{tab:stage1_and_stage2_only_fine_tune} show that among the multilingual models—InternVL2.5, Qwen2.5-VL (2B and 7B), and MiniCPM-V—InternVL2.5 consistently underperforms, with lexical scores averaging 20\% lower and precision/recall scores averaging over 7\% lower across both the findings and impressions generation stages. In contrast, Qwen2.5-VL-7B stands out as the top performer among multilingual models and across all models evaluated, consistently achieving over 60\% across all metrics in both stages. This highlights the advantages of its architectural design and the substantial Vietnamese data it was trained on, enabling stronger image-text alignment, especially in medical domains such as X-ray interpretation. However, similar to monolingual models, multilingual models also exhibit a decline in performance during the impression generation stage. 
 

\textbf{Multi-turn Generation:} The results in Table \ref{tab:stage3-maintable} indicate that most models exhibited a slight decline in performance when tasked with generating both findings and impressions in a multi-turn setting, as observed with models such as Vintern-v3.5 and InternVL2.5. In contrast, larger models like Qwen2.5-VL-7B and MiniCPM-V (8B) demonstrated notable improvements across all evaluation metrics, including lexical quality, precision, and recall. For example, Qwen2.5-VL-7B achieved a substantial boost in impressions generation, with average lexical scores increasing by more than 20\%, alongside expected gains in factual accuracy by more than 29\%, both precision and recall metrics. These findings suggest that multi-turn training more accurately mirrors the diagnostic reasoning process of radiologists, where findings are first described before clinical conclusions are drawn, and underscore the robustness of larger models when fine-tuned on a comprehensive Vietnamese instruction dataset. 

\begin{table*}[t]
\centering
\resizebox{\textwidth}{!}{%

\begin{tabular}{|cc|cccccc|cccccc|}
\hline
\multicolumn{2}{|l|}{\multirow{3}{*}{\textbf{}}} & \multicolumn{6}{c|}{\textbf{Dev}} & \multicolumn{6}{c|}{\textbf{Test}} \\ \cline{3-14} 

\multicolumn{2}{|l|}{} & \multicolumn{3}{c|}{\textbf{Rouge}} & \multicolumn{1}{c|}{\multirow{2}{*}{\textbf{Bleu}}} & \multicolumn{1}{c|}{\multirow{2}{*}{\textbf{Precision}}} & \multirow{2}{*}{\textbf{Recall}} & \multicolumn{3}{c|}{\textbf{Rouge}} & \multicolumn{1}{c|}{\multirow{2}{*}{\textbf{Bleu}}} & \multicolumn{1}{c|}{\multirow{2}{*}{\textbf{Precision}}} & \multirow{2}{*}{\textbf{Recall}} \\ \cline{3-5} \cline{9-11}

\multicolumn{2}{|l|}{} & \multicolumn{1}{c|}{\textbf{1}} & \multicolumn{1}{c|}{\textbf{2}} & \multicolumn{1}{c|}{\textbf{L}} & \multicolumn{1}{c|}{} & \multicolumn{1}{c|}{} &  & \multicolumn{1}{c|}{\textbf{1}} & \multicolumn{1}{c|}{\textbf{2}} & \multicolumn{1}{c|}{\textbf{L}} & \multicolumn{1}{c|}{} & \multicolumn{1}{c|}{} &  \\ \hline

\multicolumn{14}{|l|}{\textbf{Stage 1 - Findings Generation}}
\\ \hline

\multicolumn{2}{|c|}{Gemini} & \multicolumn{1}{c|}{44.07} & \multicolumn{1}{c|}{34.01} & \multicolumn{1}{c|}{34.49} & \multicolumn{1}{c|}{22.79} & \multicolumn{1}{c|}{12.10} & 11.20 & \multicolumn{1}{c|}{62.51} & \multicolumn{1}{c|}{35.12} & \multicolumn{1}{c|}{44.12} & \multicolumn{1}{c|}{32.31} & \multicolumn{1}{c|}{10.20} & 11.10 \\ 

\multicolumn{2}{|c|}{GPT-4v} & \multicolumn{1}{c|}{47.79} & \multicolumn{1}{c|}{17.57} & \multicolumn{1}{c|}{29.40} & \multicolumn{1}{c|}{11.54} & \multicolumn{1}{c|}{0.27} & 0.34 & \multicolumn{1}{c|}{46.51} & \multicolumn{1}{c|}{18.54} & \multicolumn{1}{c|}{31.02} & \multicolumn{1}{c|}{15.21} & \multicolumn{1}{c|}{0.51} & 0.41 \\ 

\multicolumn{2}{|c|}{Qwen2.5VL-7B} & \multicolumn{1}{c|}{\textbf{84.09}} & \multicolumn{1}{c|}{\textbf{76.77}} & \multicolumn{1}{c|}{\textbf{81.11}} & \multicolumn{1}{c|}{\textbf{70.11}} & \multicolumn{1}{c|}{\textbf{68.91}} & \textbf{69.94} & \multicolumn{1}{c|}{\textbf{84.30}} & \multicolumn{1}{c|}{\textbf{76.10}} & \multicolumn{1}{c|}{\textbf{81.21}} & \multicolumn{1}{c|}{\textbf{71.22}} & \multicolumn{1}{c|}{\textbf{70.51}} & \textbf{70.21} \\ \hline

\multicolumn{14}{|l|}{\textbf{Stage 2 - Impression Generation}}
\\ \hline

\multicolumn{2}{|c|}{Gemini} & \multicolumn{1}{c|}{39.22} & \multicolumn{1}{c|}{14.16} & \multicolumn{1}{c|}{26.72} & \multicolumn{1}{c|}{30.76} & \multicolumn{1}{c|}{0.91} & 0.75 & \multicolumn{1}{c|}{38.75} & \multicolumn{1}{c|}{15.13} & \multicolumn{1}{c|}{27.15} & \multicolumn{1}{c|}{31.54} & \multicolumn{1}{c|}{0.83} & 0.79 \\ 

\multicolumn{2}{|c|}{GPT-4v} & \multicolumn{1}{c|}{35.83} & \multicolumn{1}{c|}{8.39} & \multicolumn{1}{c|}{23.35} & \multicolumn{1}{c|}{11.66} & \multicolumn{1}{c|}{0.01} & 0.02 & \multicolumn{1}{c|}{35.54} & \multicolumn{1}{c|}{15.12} & \multicolumn{1}{c|}{25.41} & \multicolumn{1}{c|}{12.55} & \multicolumn{1}{c|}{0.02} & 0.01 \\

\multicolumn{2}{|c|}{Qwen2.5VL-7B} & \multicolumn{1}{c|}{\textbf{74.17}} & \multicolumn{1}{c|}{\textbf{65.75}} & \multicolumn{1}{c|}{\textbf{71.81}} & \multicolumn{1}{c|}{\textbf{60.11}} & \multicolumn{1}{c|}{\textbf{60.58}} & \textbf{61.94} & \multicolumn{1}{c|}{\textbf{73.89}} & \multicolumn{1}{c|}{\textbf{64.66}} & \multicolumn{1}{c|}{\textbf{71.11}} & \multicolumn{1}{c|}{\textbf{59.56}} & \multicolumn{1}{c|}{\textbf{61.25}} & \textbf{62.14} \\ \hline

\multicolumn{14}{|l|}{\textbf{Stage 3 - Multi-turn Generation}}
\\ \hline

\multicolumn{1}{|c|}{\multirow{3}{*}{\textbf{Findings}}} & Gemini & \multicolumn{1}{c|}{60.97} & \multicolumn{1}{c|}{21.67} & \multicolumn{1}{c|}{35.99} & \multicolumn{1}{c|}{41.32} & \multicolumn{1}{c|}{0.37} & 0.12 & \multicolumn{1}{c|}{61.25} & \multicolumn{1}{c|}{22.15} & \multicolumn{1}{c|}{33.25} & \multicolumn{1}{c|}{40.12} & \multicolumn{1}{c|}{0.38} & 0.15 \\ 

\multicolumn{1}{|c|}{} & GPT-4v & \multicolumn{1}{c|}{47.63} & \multicolumn{1}{c|}{13.50} & \multicolumn{1}{c|}{28.16} & \multicolumn{1}{c|}{34.52} & \multicolumn{1}{c|}{0.20} & 0.01 & \multicolumn{1}{c|}{46.52} & \multicolumn{1}{c|}{11.25} & \multicolumn{1}{c|}{25.41} & \multicolumn{1}{c|}{31.52} & \multicolumn{1}{c|}{0.25} & 0.01 \\ 

\multicolumn{1}{|c|}{} & Qwen2.5VL-7B & \multicolumn{1}{c|}{\textbf{84.81}} & \multicolumn{1}{c|}{\textbf{76.97}} & \multicolumn{1}{c|}{\textbf{80.47}} & \multicolumn{1}{c|}{\textbf{71.35}} & \multicolumn{1}{c|}{\textbf{70.34}} & \textbf{72.28} & \multicolumn{1}{c|}{\textbf{84.40}} & \multicolumn{1}{c|}{\textbf{75.90}} & \multicolumn{1}{c|}{\textbf{79.60}} & \multicolumn{1}{c|}{\textbf{69.85}} & \multicolumn{1}{c|}{\textbf{69.78}} & \textbf{70.60} \\ \cline{1-14} 

\multicolumn{1}{|c|}{\multirow{3}{*}{\textbf{Impression}}} & Gemini & \multicolumn{1}{c|}{31.96} & \multicolumn{1}{c|}{10.03} & \multicolumn{1}{c|}{25.69} & \multicolumn{1}{c|}{42.21} & \multicolumn{1}{c|}{0.33} & 0.42 & \multicolumn{1}{c|}{32.21} & \multicolumn{1}{c|}{9.51} & \multicolumn{1}{c|}{21.54} & \multicolumn{1}{c|}{42.51} & \multicolumn{1}{c|}{0.41} & 0.35 \\ 

\multicolumn{1}{|c|}{} & GPT-4v & \multicolumn{1}{c|}{31.75} & \multicolumn{1}{c|}{6.94} & \multicolumn{1}{c|}{18.59} & \multicolumn{1}{c|}{21.51} & \multicolumn{1}{c|}{0.30} & 0.01 & \multicolumn{1}{c|}{32.15} & \multicolumn{1}{c|}{11.25} & \multicolumn{1}{c|}{20.12} & \multicolumn{1}{c|}{20.52} & \multicolumn{1}{c|}{0.21} & 0.02 \\ 

\multicolumn{1}{|c|}{} & Qwen2.5VL-7B & \multicolumn{1}{c|}{\textbf{95.93}} & \multicolumn{1}{c|}{\textbf{94.34}} & \multicolumn{1}{c|}{\textbf{95.08}} & \multicolumn{1}{c|}{\textbf{92.04}} & \multicolumn{1}{c|}{\textbf{92.14}} & \textbf{92.11} & \multicolumn{1}{c|}{\textbf{95.20}} & \multicolumn{1}{c|}{\textbf{93.80}} & \multicolumn{1}{c|}{\textbf{94.68}} & \multicolumn{1}{c|}{\textbf{89.75}} & \multicolumn{1}{c|}{\textbf{89.95}} & \textbf{90.88} \\ \hline
\end{tabular}
}
\caption{Comparison of Qwen2.5VL-7B performance (\%) with Gemini and GPT-4V (o4 multimodal version) across three stages.}
\label{tab:qwen-Gemini-GPT}
\end{table*}

\textbf{Compare with Gemini and GPT-4v:} In addition to comparing against open-source VLMs, we also evaluated Gemini \citep{gemini} and GPT-4v (o4 multimodal version) \citep{gpt-4o-v} using the same input format described in Section \ref{subsec:Experiment_Setup}. The generated outputs were assessed using the same evaluation metrics outlined in Section \ref{subsec:Evaluation_Metrics} and compared to our best-performing fine-tuned model, Qwen2.5-VL-7B. As shown in Table \ref{tab:qwen-Gemini-GPT}, while Gemini and GPT-4v occasionally produce outputs resembling radiologist-style findings and impressions, their overall precision and recall remain low, often failing to generate any accurate information. 
Furthermore, we conducted a manual evaluation of the generative outputs of the models. In this process, we categorized the generated information into three main types: correct information, incorrect information, and redundant information. The results reveal that while Gemini can sometimes produce correct statements, they are often overshadowed by a large amount of unnecessary content. Moreover, GPT-4v occasionally refuses to generate outputs based on our provided inputs, likely due to its built-in constraints related to clinical accuracy. In contrast, Qwen2.5-VL-7B consistently delivers more complete and accurate responses, highlighting the potential of open-source models not only for our medical VLM task but also for healthcare applications more broadly.

\section{Conclusion}
In this study, we introduce a novel dataset named ViX-Ray, collected from radiological findings and diagnostic impressions written by physicians at the Vietnamese Military Hospital 175, based on chest X-ray images of Vietnamese patients. We conduct a detailed analysis of the dataset, including body part and diagnosis frequency distributions, to gain a deeper understanding of the patterns present in clinical findings and impressions. For experimentation, we fine-tune state-of-the-art vision-language models (VLMs), ranging from multilingual to Vietnamese monolingual models, on our dataset. We also benchmark their performance against proprietary models such as GPT-4v and Gemini to provide a comprehensive evaluation of current VLMs on our data. Experimental results show that Qwen2.5-VL-7B consistently outperforms other models across multiple evaluation metrics. 

Despite its contributions, the ViX-Ray dataset still exhibits limitations in both scale and diversity. In terms of size, it contains fewer samples—ranging from one-half to one-twentieth—compared to other datasets such as ChestX-ray8 and VinDr-CXR. This relatively small size limits the representation of diverse pathological cases involving the heart, lungs, and other thoracic regions. Future work will focus on expanding the dataset in both scale and variety to support more comprehensive research on X-ray-based medical diagnosis in Vietnamese and contribute to broader efforts in developing Vietnamese medical AI.

\section{Limitation}
\textbf{Limitations of the ViX-Ray dataset:} Due to the nature of the dataset, which is constructed from diagnostic reports written by radiologists, its use is inherently limited to specific tasks. The dataset consists solely of written medical impressions, lacking detailed annotations about the exact anatomical locations of abnormalities (such as bone, liver, or heart regions) on X-ray images of the patient.
As a result, models trained on this dataset can only provide general descriptions without explicitly localizing findings on the image, reducing the overall challenge for current VLMs. 
Furthermore, in terms of data coverage, the dataset remains relatively small compared to similar resources in other languages, thereby limiting the breadth of medical knowledge available to the broader research community.

\textbf{Limitations in Experiments:} Our experiments were limited to open-source Vietnamese VLMs, and we only tested models with parameter sizes below 7B. 
This inevitably restricted the performance potential of some models on our dataset and excluded an evaluation of closed-source models, such as GPT or Gemini, which may offer stronger capabilities. 
Moreover, we used a standardized instruction prompt throughout all three fine-tuning stages to maintain consistency in evaluating model performance. 
While this approach ensured fair comparisons, it also meant we did not explore alternative prompt designs or prompting strategies that have been shown in other studies to significantly enhance the model effectiveness.




\section{Bibliographical References}

\bibliographystyle{lrec2026-natbib}
\bibliography{lrec2026-example}

\newpage
\onecolumn

\appendix
\section{Prompt}
\label{sec:appendixA}
\subsection{Stage 1 - Findings Generation}
\label{sec:appendixA1}
We provide patient information, including gender, age, and the X-ray view type, along with the X-ray image of the patient. An example of the input used in Stage 1 - findings generation is illustrated in Table \ref{tab:stage1_input_example}.
\begin{table*}[t]
    \centering
    \begin{tabular}{|c|p{0.9\linewidth}|}
    \hline
    & \\
         \multirow{25}{*}{\textbf{\shortstack{Stage 1 -\\Findings\\Generation}}} & \textbf{"Role"}: "System" $-$ \textbf{"Content"}: "Bạn là trợ lý bác sĩ (\textit{You are a medical assistant})"\\ 
        &  \textbf{"Role"}: "User" $-$ \textbf{"Content"}: "Ảnh chụp X-ray PA (Chụp Xquang tim phổi thẳng) bệnh nhân nam, 72 tuổi. Cho biết bệnh nhân bị gì?" 
        
        $($\textit{X-ray PA image (Posteroanterior chest X-ray) of a 72-year-old male patient. Please indicate what condition the patient has}$)$ \\
        & \textbf{"Role"}: "User" $-$ \textbf{"Content"}: \\
        & \multicolumn{1}{c|}{\includegraphics[width=0.3 \linewidth]{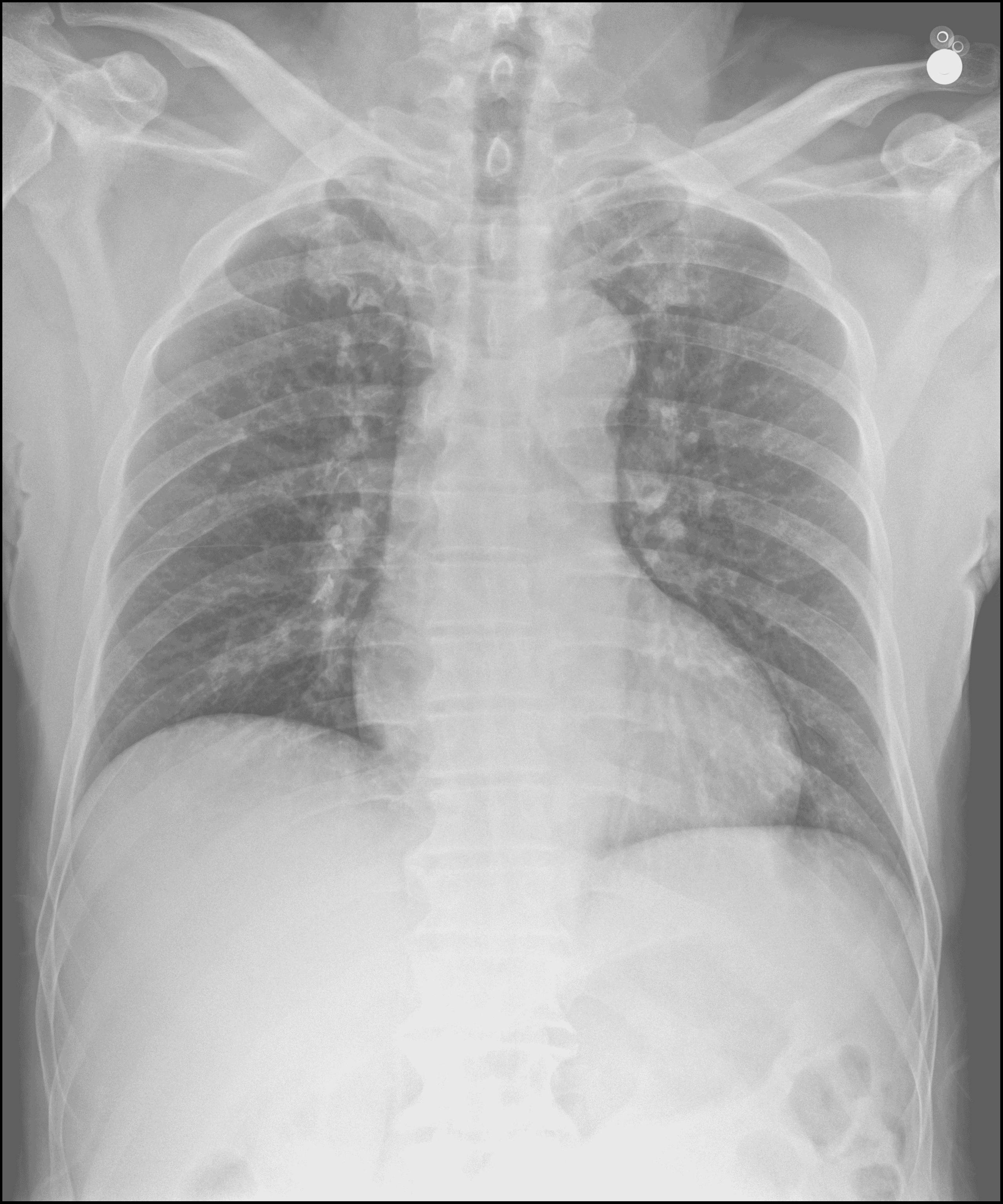}}\\
          & \textbf{"Ground Truth"}: "Phổi hai bên kém sáng, mờ kính rải rác, rốn phổi tăng đậm nhẹ. Bờ vòm hoành hai bên đều, hai góc sườn hoành nhọn. Bóng mờ tim không to. Vôi hóa thành quai động mạch chủ."
          
          (\textit{The lungs are bilaterally hypolucent with scattered ground-glass opacities. Mild hilar prominence is noted. The diaphragmatic contours are smooth, and both costophrenic angles are sharp. The cardiac silhouette is within normal limits. Calcifications are seen in the wall of the aortic arch.})\\ \hline
    \end{tabular}
    \caption{An example of Stage 1 input for findings generation}
    \label{tab:stage1_input_example}
\end{table*}
\begin{table}[b]
    \centering
    \begin{tabular}{|l:c|}
    \hline
    \multicolumn{2}{|c|}{\textbf{Hyper-parameters}} 
    \\ \hline
         Epoch&	20	\\
Learning rate&	3.00E-05 \\	
Batch size&	8	\\
Gradient accumulate&	4	\\
Optim&	AdamW	\\
Loss&	Cross entropy	\\
\hline
    \end{tabular}
    \caption{Summary of Hyperparameters Used During Fine-tuning}
    \label{tab:hyper-parameters}
\end{table}

\subsection{Stage 2 - Findings Generation}
\label{sec:appendixA2}
Similar to the input used for Stage 1 — findings generation during VLM fine-tuning, we also provide the necessary patient information and X-ray image, as illustrated in Table \ref{tab:stage2_input_example}.

\begin{table*}[]
    \centering
    \begin{tabular}{|c|p{0.9\linewidth}|}
    \hline
    & \\
         \multirow{25}{*}{\textbf{\shortstack{Stage 2 -\\Impressions\\Generation}}} & \textbf{"Role"}: "System" $-$ \textbf{"Content"}: "Bạn là trợ lý bác sĩ (\textit{You are a medical assistant})"\\ 
        &  \textbf{"Role"}: "User" $-$ \textbf{"Content"}: "Ảnh chụp X-ray PA (Chụp Xquang tim phổi thẳng) bệnh nhân nữ, 68 tuổi. Cho biết bệnh nhân bị gì?" 
        
        $($\textit{X-ray PA image (Posteroanterior chest X-ray) of a 68-year-old female patient. Please indicate what condition the patient has.}$)$ \\
        & \textbf{"Role"}: "User" $-$ \textbf{"Content"}: \\
        & \multicolumn{1}{c|}{\includegraphics[width=0.3 \linewidth]{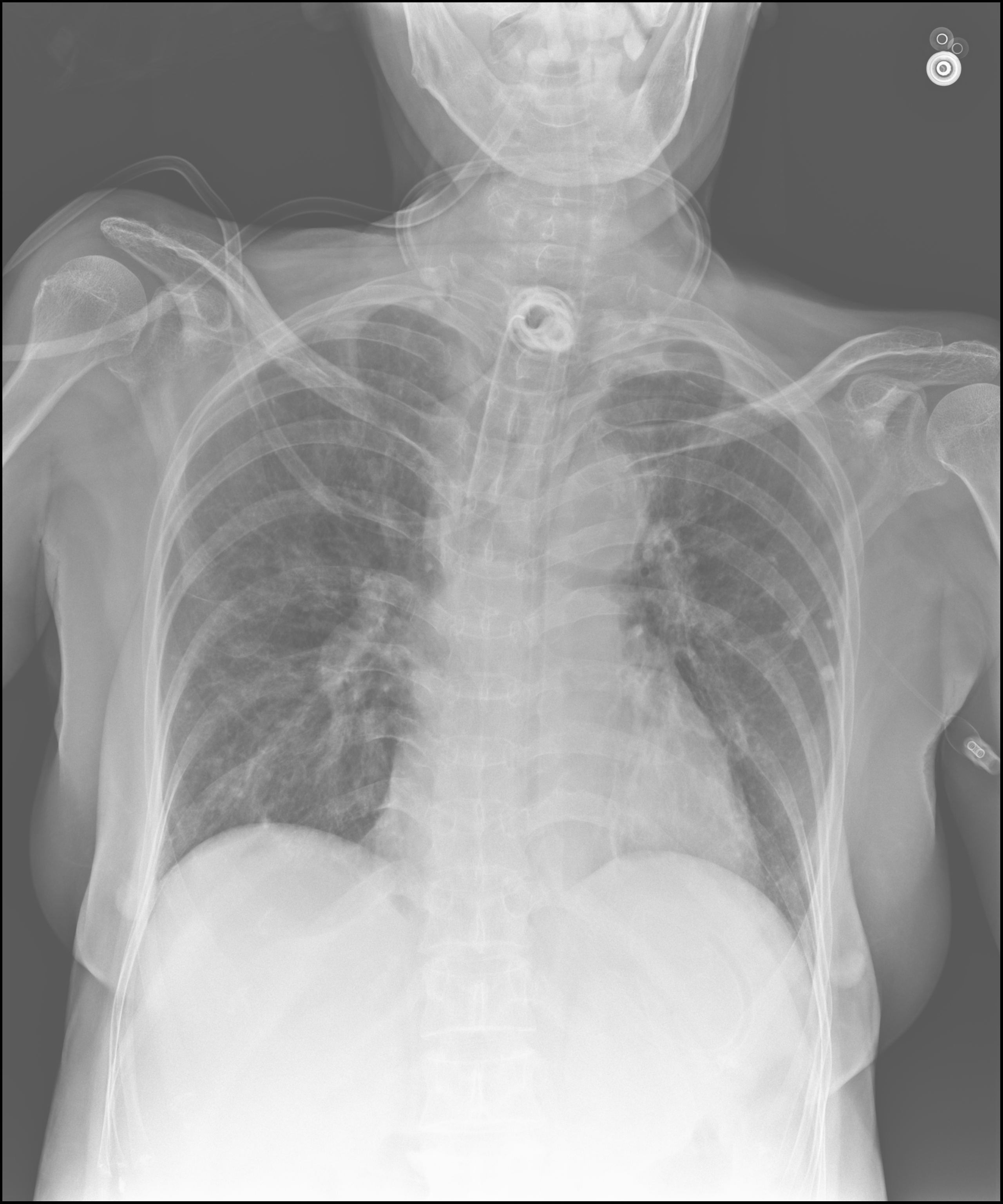}}\\
          & \textbf{"Ground Truth"}: "Hình ảnh nghĩ nhiều đến tổn thương phổi kẽ. Vôi hóa thùy trên phổi trái"
          
          (\textit{The image is highly suggestive of interstitial lung lesions. Calcification in the upper lobe of the left lung.})\\ \hline
    \end{tabular}
    \caption{An example of Stage 2 input for impressions generation.}
    \label{tab:stage2_input_example}
\end{table*}

\subsection{Stage 3 - Multi-turn Generation}
\label{sec:appendixA3}
We illustrate the input for Stage 3 — multi-turn generation in Table \ref{tab:stage3_input_example}. In this setup, we provide patient information, including age, gender, and the X-ray image. Based on this input, the model is first required to describe the condition of the patient (findings), and then to generate diagnostic conclusions (impressions) from the described information.
\begin{table}[]
    \centering
    \begin{tabular}{|c|p{0.9\linewidth}|}
    \hline
    & \\
         \multirow{31}{*}{\textbf{\shortstack{Stage 3 -\\Multi-turn\\Generation}}} & \textbf{"Role"}: "System" $-$ \textbf{"Content"}: "Bạn là trợ lý bác sĩ (\textit{You are a medical assistant})"\\ 
         
        &  \textbf{"Role"}: "User" $-$ \textbf{"Content"}: "Ảnh chụp X-ray PA (Chụp Xquang tim phổi thẳng) bệnh nhân nam, 56 tuổi. Cho biết bệnh nhân bị gì?" 
        
        $($\textit{X-ray PA image (Posteroanterior chest X-ray) of a 56-year-old male patient. Please indicate what condition the patient has.}$)$ \\
        & \textbf{"Role"}: "User" $-$ \textbf{"Content"}: \\
        & \multicolumn{1}{c|}{\includegraphics[width=0.3 \linewidth]{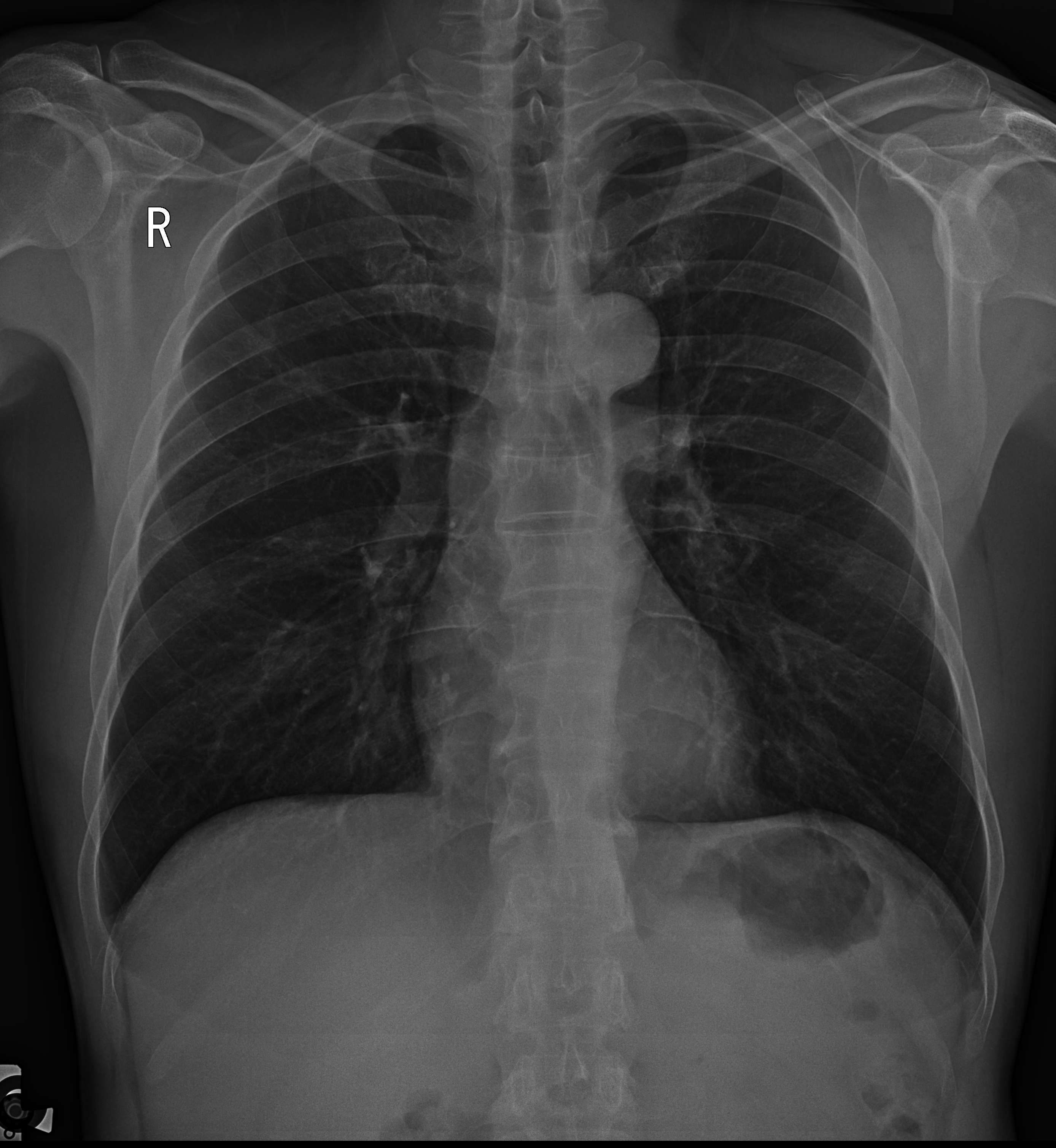}}\\
          &  \textbf{"Role"}: "Assistant" $-$ \textbf{"Content"}: "Lồng ngực cân đối, không thấy bất thường xương. Phổi hai bên sáng đều. Rốn phổi hai bên không thấy bất thường. Bờ vòm hoành hai bên đều. Góc sườn hoành hai bên nhọn. Bóng tim không to. Quai động mạch chủ vồng." 
          
          (\textit{The chest appears symmetrical with no bony abnormalities. Lungs are bilaterally clear. Hilar regions are unremarkable. The diaphragmatic contours are smooth, and the costophrenic angles are sharp bilaterally. The cardiac silhouette is within normal limits. The aortic arch appears prominent.})\\ 
          &  \textbf{"Role"}: "User" $-$ \textbf{"Content"}: "Kết luận bệnh nhân bị gì?" 
          
          (\textit{What is the diagnosis of the patient?})\\
          &  \textbf{"Role"}: "User" $-$ \textbf{"Content"}: <image> \\
          
          &  \textbf{"Role"}: "User" $-$ \textbf{"Content"}: "Hình ảnh Quai động mạch chủ vồng." 
          
          (\textit{The image shows a prominent aortic arch.})\\ \hline

    \end{tabular}
    \caption{An example of Stage 3 input for multi-turn generation.}
    \label{tab:stage3_input_example}
\end{table}

\section{Training Hyperparameters}
\label{sec:appendixHyperParams}
We summarize the hyperparameters in Table \ref{tab:hyper-parameters}. The 1B models were trained on two RTX 3090 24GB GPUs, while the remaining models were trained on seven RTX 5090 GPUs.

\section{Training Result}
\label{sec:appendixTrainingResult}
In this section, we present the results of various Visual Language Models (VLMs) on our ViX-Ray dataset, both before and after the fine-tuning process.

\subsection{Findings Generation Training Result}
\label{appendix:Training_Result_Stage1}
As shown in Table \ref{tab:training_result_stage1}, fine-tuning significantly improved the performance of the VLMs. This enhancement was particularly notable in the accuracy of generated information, as evidenced by substantial gains in both precision and recall.

\begin{table}[]
\centering
\resizebox{\textwidth}{!}{%
\begin{tabular}{|clclclclclccclclclclcc|}
\hline
\multicolumn{2}{|l|}{\multirow{3}{*}{}} & \multicolumn{10}{c|}{\textbf{Dev}} & \multicolumn{10}{c|}{\textbf{Test}} \\ \cline{3-22} 
\multicolumn{2}{|l|}{} & \multicolumn{6}{c|}{\textbf{Rouge}} & \multicolumn{2}{c|}{\multirow{2}{*}{\textbf{Bleu}}} & \multicolumn{1}{c|}{\multirow{2}{*}{\textbf{Precision}}} & \multicolumn{1}{c|}{\multirow{2}{*}{\textbf{Recall}}} & \multicolumn{6}{c|}{\textbf{Rouge}} & \multicolumn{2}{c|}{\multirow{2}{*}{\textbf{Bleu}}} & \multicolumn{1}{c|}{\multirow{2}{*}{\textbf{Precision}}} & \multirow{2}{*}{\textbf{Recall}} \\ \cline{3-8} \cline{13-18}
\multicolumn{2}{|l|}{} & \multicolumn{2}{c|}{\textbf{1}} & \multicolumn{2}{c|}{\textbf{2}} & \multicolumn{2}{c|}{\textbf{L}} & \multicolumn{2}{c|}{} & \multicolumn{1}{c|}{} & \multicolumn{1}{c|}{} & \multicolumn{2}{c|}{\textbf{1}} & \multicolumn{2}{c|}{\textbf{2}} & \multicolumn{2}{c|}{\textbf{L}} & \multicolumn{2}{c|}{} & \multicolumn{1}{c|}{} &  \\ \hline
\multicolumn{22}{|l|}{\textbf{None Fine-tune}} \\ \hline
\multicolumn{1}{|c|}{\multirow{2}{*}{\textbf{Monolingual}}} & \multicolumn{1}{l|}{Vintern-v3.5} & \multicolumn{2}{c|}{43.04} & \multicolumn{2}{c|}{11.10} & \multicolumn{2}{c|}{23.70} & \multicolumn{2}{c|}{5.49} & \multicolumn{1}{c|}{0.01} & \multicolumn{1}{c|}{0.01} & \multicolumn{2}{c|}{42.78} & \multicolumn{2}{c|}{10.89} & \multicolumn{2}{c|}{23.72} & \multicolumn{2}{c|}{5.94} & \multicolumn{1}{c|}{0.01} & 0.01 \\ 
\multicolumn{1}{|c|}{} & \multicolumn{1}{l|}{LaVy} & \multicolumn{2}{c|}{45.53} & \multicolumn{2}{c|}{\textbf{12.06}} & \multicolumn{2}{c|}{25.32} & \multicolumn{2}{c|}{\textbf{19.52}} & \multicolumn{1}{c|}{0.01} & \multicolumn{1}{c|}{0.01} & \multicolumn{2}{c|}{45.68} & \multicolumn{2}{c|}{\textbf{12.01}} & \multicolumn{2}{c|}{25.60} & \multicolumn{2}{c|}{\textbf{10.66}} & \multicolumn{1}{c|}{0.01} & 0.01 \\ \hline
\multicolumn{1}{|c|}{\multirow{4}{*}{\textbf{Multilingual}}} & \multicolumn{1}{l|}{InternVL2.5} & \multicolumn{2}{c|}{50.27} & \multicolumn{2}{c|}{10.31} & \multicolumn{2}{c|}{24.95} & \multicolumn{2}{c|}{2.87} & \multicolumn{1}{c|}{0.01} & \multicolumn{1}{c|}{0.01} & \multicolumn{2}{c|}{50.21} & \multicolumn{2}{c|}{10.31} & \multicolumn{2}{c|}{24.97} & \multicolumn{2}{c|}{3.08} & \multicolumn{1}{c|}{0.01} & 0.01 \\ 
\multicolumn{1}{|c|}{} & \multicolumn{1}{l|}{Qwen2.5VL-2B} & \multicolumn{2}{c|}{44.54} & \multicolumn{2}{c|}{10.53} & \multicolumn{2}{c|}{24.81} & \multicolumn{2}{c|}{1.10} & \multicolumn{1}{c|}{0.01} & \multicolumn{1}{c|}{0.01} & \multicolumn{2}{c|}{45.74} & \multicolumn{2}{c|}{10.81} & \multicolumn{2}{c|}{25.24} & \multicolumn{2}{c|}{1.20} & \multicolumn{1}{c|}{0.01} & 0.01 \\ 
\multicolumn{1}{|c|}{} & \multicolumn{1}{l|}{Qwen2.5VL-7B} & \multicolumn{2}{c|}{\textbf{54.31}} & \multicolumn{2}{c|}{11.17} & \multicolumn{2}{c|}{\textbf{30.35}} & \multicolumn{2}{c|}{1.20} & \multicolumn{1}{c|}{0.01} & \multicolumn{1}{c|}{0.01} & \multicolumn{2}{c|}{\textbf{54.22}} & \multicolumn{2}{c|}{11.18} & \multicolumn{2}{c|}{\textbf{30.19}} & \multicolumn{2}{c|}{1.10} & \multicolumn{1}{c|}{0.01} & 0.01 \\ 
\multicolumn{1}{|c|}{} & \multicolumn{1}{l|}{MiniCPM-V} & \multicolumn{2}{c|}{27.18} & \multicolumn{2}{c|}{5.71} & \multicolumn{2}{c|}{15.49} & \multicolumn{2}{c|}{1.49} & \multicolumn{1}{c|}{0.01} & \multicolumn{1}{c|}{0.01} & \multicolumn{2}{c|}{26.64} & \multicolumn{2}{c|}{5.76} & \multicolumn{2}{c|}{15.12} & \multicolumn{2}{c|}{1.59} & \multicolumn{1}{c|}{0.01} & 0.01 \\ \hline
\multicolumn{22}{|l|}{\textbf{Fine-tuning}} \\ \hline
\multicolumn{1}{|c|}{\multirow{2}{*}{\textbf{Monolingual}}} & \multicolumn{1}{l|}{Vintern-v3.5} & \multicolumn{2}{c|}{81.79} & \multicolumn{2}{c|}{71.71} & \multicolumn{2}{c|}{75.99} & \multicolumn{2}{c|}{61.91} & \multicolumn{1}{c|}{57.25} & \multicolumn{1}{c|}{56.74} & \multicolumn{2}{c|}{81.90} & \multicolumn{2}{c|}{72.10} & \multicolumn{2}{c|}{76.21} & \multicolumn{2}{c|}{61.39} & \multicolumn{1}{c|}{56.25} & 56.12 \\ 
\multicolumn{1}{|c|}{} & \multicolumn{1}{l|}{LaVy} & \multicolumn{2}{c|}{81.95} & \multicolumn{2}{c|}{71.08} & \multicolumn{2}{c|}{76.91} & \multicolumn{2}{c|}{69.11} & \multicolumn{1}{c|}{61.54} & \multicolumn{1}{c|}{61.12} & \multicolumn{2}{c|}{82.12} & \multicolumn{2}{c|}{71.35} & \multicolumn{2}{c|}{77.25} & \multicolumn{2}{c|}{69.58} & \multicolumn{1}{c|}{62.48} & 61.52 \\ \hline
\multicolumn{1}{|c|}{\multirow{4}{*}{\textbf{Multilingual}}} & \multicolumn{1}{l|}{InternVL2.5} & \multicolumn{2}{c|}{66.29} & \multicolumn{2}{c|}{47.18} & \multicolumn{2}{c|}{51.06} & \multicolumn{2}{c|}{62.09} & \multicolumn{1}{c|}{56.24} & \multicolumn{1}{c|}{55.12} & \multicolumn{2}{c|}{65.67} & \multicolumn{2}{c|}{46.82} & \multicolumn{2}{c|}{50.57} & \multicolumn{2}{c|}{61.94} & \multicolumn{1}{c|}{56.24} & 55.12 \\ 
\multicolumn{1}{|c|}{} & \multicolumn{1}{l|}{Qwen2.5VL-2B} & \multicolumn{2}{c|}{83.40} & \multicolumn{2}{c|}{75.25} & \multicolumn{2}{c|}{79.03} & \multicolumn{2}{c|}{69.42} & \multicolumn{1}{c|}{61.42} & \multicolumn{1}{c|}{60.24} & \multicolumn{2}{c|}{83.87} & \multicolumn{2}{c|}{75.56} & \multicolumn{2}{c|}{79.27} & \multicolumn{2}{c|}{70.21} & \multicolumn{1}{c|}{62.21} & 61.45 \\ 
\multicolumn{1}{|c|}{} & \multicolumn{1}{l|}{Qwen2.5VL-7B} & \multicolumn{2}{c|}{\textbf{84.09}} & \multicolumn{2}{c|}{\textbf{76.77}} & \multicolumn{2}{c|}{\textbf{81.11}} & \multicolumn{2}{c|}{\textbf{70.11}} & \multicolumn{1}{c|}{\textbf{68.91}} & \multicolumn{1}{c|}{\textbf{69.94}} & \multicolumn{2}{c|}{\textbf{84.30}} & \multicolumn{2}{c|}{\textbf{76.10}} & \multicolumn{2}{c|}{\textbf{81.21}} & \multicolumn{2}{c|}{\textbf{71.22}} & \multicolumn{1}{c|}{\textbf{70.51}} & \textbf{70.21} \\ 
\multicolumn{1}{|c|}{} & \multicolumn{1}{l|}{MiniCPM-V} & \multicolumn{2}{c|}{83.21} & \multicolumn{2}{c|}{73.21} & \multicolumn{2}{c|}{79.98} & \multicolumn{2}{c|}{69.24} & \multicolumn{1}{c|}{67.56} & \multicolumn{1}{c|}{66.12} & \multicolumn{2}{c|}{83.25} & \multicolumn{2}{c|}{72.97} & \multicolumn{2}{c|}{79.81} & \multicolumn{2}{c|}{70.12} & \multicolumn{1}{c|}{68.12} & 65.25 \\ \hline
\end{tabular}%
}
\caption{Results of Stage 1 - findings generation on the ViX-Ray dataset (\%), we report the performance of models before and after the fine-tuning process.}
\label{tab:training_result_stage1}
\end{table}

\subsection{Impressions Generation Training Result}
\label{appendix:Training_Result_Stage2}
Table \ref{tab:training_result_stage2} illustrates the evaluation results of VLMs for Stage 2 - impression generation, both before and after fine-tuning. Similar to the outcomes observed in Stage 1 - findings generation, the models in Stage 2 also demonstrated a significant performance improvement, with increases across various lexical and contextual metrics. However, overall results remain suboptimal, indicating that current VLMs still face challenges in generating impressions with human-level accuracy.
\begin{table}[]
\centering
\resizebox{\textwidth}{!}{%
\begin{tabular}{|clclclclclccclclclclcc|}
\hline
\multicolumn{2}{|l|}{\multirow{3}{*}{}} & \multicolumn{10}{c|}{\textbf{Dev}} & \multicolumn{10}{c|}{\textbf{Test}} \\ \cline{3-22} 
\multicolumn{2}{|l|}{} & \multicolumn{6}{c|}{\textbf{Rouge}} & \multicolumn{2}{c|}{\multirow{2}{*}{\textbf{Bleu}}} & \multicolumn{1}{c|}{\multirow{2}{*}{\textbf{Precision}}} & \multicolumn{1}{c|}{\multirow{2}{*}{\textbf{Recall}}} & \multicolumn{6}{c|}{\textbf{Rouge}} & \multicolumn{2}{c|}{\multirow{2}{*}{\textbf{Bleu}}} & \multicolumn{1}{c|}{\multirow{2}{*}{\textbf{Precision}}} & \multirow{2}{*}{\textbf{Recall}} \\ \cline{3-8} \cline{13-18}
\multicolumn{2}{|l|}{} & \multicolumn{2}{c|}{\textbf{1}} & \multicolumn{2}{c|}{\textbf{2}} & \multicolumn{2}{c|}{\textbf{L}} & \multicolumn{2}{c|}{} & \multicolumn{1}{c|}{} & \multicolumn{1}{c|}{} & \multicolumn{2}{c|}{\textbf{1}} & \multicolumn{2}{c|}{\textbf{2}} & \multicolumn{2}{c|}{\textbf{L}} & \multicolumn{2}{c|}{} & \multicolumn{1}{c|}{} &  \\ \hline
\multicolumn{22}{|l|}{\textbf{None Fine-tune}} \\ \hline
\multicolumn{1}{|c|}{\multirow{2}{*}{Monolingual}} & \multicolumn{1}{l|}{Vintern-v3.5} & \multicolumn{2}{c|}{25.39} & \multicolumn{2}{c|}{8.07} & \multicolumn{2}{c|}{18.53} & \multicolumn{2}{c|}{5.44} & \multicolumn{1}{c|}{0.01} & \multicolumn{1}{c|}{0.01} & \multicolumn{2}{c|}{25.27} & \multicolumn{2}{c|}{8.00} & \multicolumn{2}{c|}{18.35} & \multicolumn{2}{c|}{4.67} & \multicolumn{1}{c|}{0.01} & 0.01 \\ 
\multicolumn{1}{|c|}{} & \multicolumn{1}{l|}{LaVy} & \multicolumn{2}{c|}{27.94} & \multicolumn{2}{c|}{\textbf{8.62}} & \multicolumn{2}{c|}{\textbf{20.39}} & \multicolumn{2}{c|}{\textbf{11.81}} & \multicolumn{1}{c|}{0.01} & \multicolumn{1}{c|}{0.01} & \multicolumn{2}{c|}{27.69} & \multicolumn{2}{c|}{\textbf{8.50}} & \multicolumn{2}{c|}{\textbf{20.08}} & \multicolumn{2}{c|}{\textbf{12.77}} & \multicolumn{1}{c|}{0.01} & 0.01 \\ \hline
\multicolumn{1}{|c|}{\multirow{4}{*}{Multilingual}} & \multicolumn{1}{l|}{InternVL2.5} & \multicolumn{2}{c|}{26.41} & \multicolumn{2}{c|}{7.33} & \multicolumn{2}{c|}{18.59} & \multicolumn{2}{c|}{3.84} & \multicolumn{1}{c|}{0.01} & \multicolumn{1}{c|}{0.01} & \multicolumn{2}{c|}{26.76} & \multicolumn{2}{c|}{7.22} & \multicolumn{2}{c|}{18.44} & \multicolumn{2}{c|}{3.28} & \multicolumn{1}{c|}{0.01} & 0.01 \\ 
\multicolumn{1}{|c|}{} & \multicolumn{1}{l|}{Qwen2.5VL-2B} & \multicolumn{2}{c|}{20.74} & \multicolumn{2}{c|}{5.90} & \multicolumn{2}{c|}{14.37} & \multicolumn{2}{c|}{1.10} & \multicolumn{1}{c|}{0.01} & \multicolumn{1}{c|}{0.01} & \multicolumn{2}{c|}{21.45} & \multicolumn{2}{c|}{5.99} & \multicolumn{2}{c|}{14.65} & \multicolumn{2}{c|}{1.20} & \multicolumn{1}{c|}{0.01} & 0.01 \\ 
\multicolumn{1}{|c|}{} & \multicolumn{1}{l|}{Qwen2.5VL-7B} & \multicolumn{2}{c|}{\textbf{31.77}} & \multicolumn{2}{c|}{7.80} & \multicolumn{2}{c|}{18.06} & \multicolumn{2}{c|}{1.20} & \multicolumn{1}{c|}{0.01} & \multicolumn{1}{c|}{0.01} & \multicolumn{2}{c|}{\textbf{31.74}} & \multicolumn{2}{c|}{8.00} & \multicolumn{2}{c|}{18.06} & \multicolumn{2}{c|}{1.10} & \multicolumn{1}{c|}{0.01} & 0.01 \\ 
\multicolumn{1}{|c|}{} & \multicolumn{1}{l|}{MiniCPM-V} & \multicolumn{2}{c|}{14.82} & \multicolumn{2}{c|}{3.72} & \multicolumn{2}{c|}{10.93} & \multicolumn{2}{c|}{1.04} & \multicolumn{1}{c|}{0.01} & \multicolumn{1}{c|}{0.01} & \multicolumn{2}{c|}{15.81} & \multicolumn{2}{c|}{3.84} & \multicolumn{2}{c|}{11.25} & \multicolumn{2}{c|}{1.20} & \multicolumn{1}{c|}{0.01} & 0.01 \\ \hline
\multicolumn{22}{|l|}{\textbf{Fine-tuning}} \\ \hline
\multicolumn{1}{|c|}{\multirow{2}{*}{Monolingual}} & \multicolumn{1}{l|}{Vintern-v3.5} & \multicolumn{2}{c|}{62.82} & \multicolumn{2}{c|}{52.16} & \multicolumn{2}{c|}{57.98} & \multicolumn{2}{c|}{51.13} & \multicolumn{1}{c|}{56.74} & \multicolumn{1}{c|}{49.06} & \multicolumn{2}{c|}{60.83} & \multicolumn{2}{c|}{50.85} & \multicolumn{2}{c|}{56.25} & \multicolumn{2}{c|}{51.91} & \multicolumn{1}{c|}{56.41} & 50.14 \\ 
\multicolumn{1}{|c|}{} & \multicolumn{1}{l|}{LaVy} & \multicolumn{2}{c|}{70.14} & \multicolumn{2}{c|}{62.15} & \multicolumn{2}{c|}{68.45} & \multicolumn{2}{c|}{57.02} & \multicolumn{1}{c|}{57.49} & \multicolumn{1}{c|}{51.21} & \multicolumn{2}{c|}{70.25} & \multicolumn{2}{c|}{61.58} & \multicolumn{2}{c|}{67.78} & \multicolumn{2}{c|}{57.45} & \multicolumn{1}{c|}{57.65} & 52.74 \\ \hline
\multicolumn{1}{|c|}{\multirow{4}{*}{Multilingual}} & \multicolumn{1}{l|}{InternVL2.5} & \multicolumn{2}{c|}{61.65} & \multicolumn{2}{c|}{41.08} & \multicolumn{2}{c|}{54.22} & \multicolumn{2}{c|}{52.10} & \multicolumn{1}{c|}{51.21} & \multicolumn{1}{c|}{48.51} & \multicolumn{2}{c|}{69.05} & \multicolumn{2}{c|}{42.79} & \multicolumn{2}{c|}{56.22} & \multicolumn{2}{c|}{52.02} & \multicolumn{1}{c|}{52.35} & 49.32 \\ 
\multicolumn{1}{|c|}{} & \multicolumn{1}{l|}{Qwen2.5VL-2B} & \multicolumn{2}{c|}{73.26} & \multicolumn{2}{c|}{64.31} & \multicolumn{2}{c|}{70.81} & \multicolumn{2}{c|}{57.11} & \multicolumn{1}{c|}{54.38} & \multicolumn{1}{c|}{53.73} & \multicolumn{2}{c|}{71.75} & \multicolumn{2}{c|}{62.48} & \multicolumn{2}{c|}{69.27} & \multicolumn{2}{c|}{57.39} & \multicolumn{1}{c|}{55.01} & 53.12 \\ 
\multicolumn{1}{|c|}{} & \multicolumn{1}{l|}{Qwen2.5VL-7B} & \multicolumn{2}{c|}{\textbf{74.17}} & \multicolumn{2}{c|}{\textbf{65.75}} & \multicolumn{2}{c|}{\textbf{71.81}} & \multicolumn{2}{c|}{\textbf{60.11}} & \multicolumn{1}{c|}{\textbf{60.58}} & \multicolumn{1}{c|}{\textbf{61.94}} & \multicolumn{2}{c|}{\textbf{73.89}} & \multicolumn{2}{c|}{\textbf{64.66}} & \multicolumn{2}{c|}{\textbf{71.11}} & \multicolumn{2}{c|}{\textbf{59.56}} & \multicolumn{1}{c|}{\textbf{61.25}} & \textbf{62.14} \\ 
\multicolumn{1}{|c|}{} & \multicolumn{1}{l|}{MiniCPM-V} & \multicolumn{2}{c|}{71.12} & \multicolumn{2}{c|}{63.31} & \multicolumn{2}{c|}{69.28} & \multicolumn{2}{c|}{58.12} & \multicolumn{1}{c|}{58.49} & \multicolumn{1}{c|}{50.84} & \multicolumn{2}{c|}{70.16} & \multicolumn{2}{c|}{61.97} & \multicolumn{2}{c|}{68.25} & \multicolumn{2}{c|}{57.25} & \multicolumn{1}{c|}{57.48} & 40.22 \\ \hline
\end{tabular}%
}
\caption{Results of Stage 2 - impressions generation on the ViX-Ray dataset (\%), we report the performance of models before and after the fine-tuning process.}
\label{tab:training_result_stage2}
\end{table}

\subsection{Multi-turn Generation Training Result}
\label{appendix:Training_Result_Stage3}
For Stage 3 - multi-turn generation, we only report the post-fine-tuning results of the VLMs. We also quantify the performance difference between Stage 3 and the other two stages. An increase in performance is highlighted with \textcolor{blue}{blue up arrow (\textuparrow)}, while a decrease is indicated by \textcolor{red}{red down arrow (\textdownarrow)}.

Our findings, illustrated in Table \ref{tab:stage3-maintable}, demonstrate that employing a multi-turn approach enhances model performance. This aligns with how doctors typically assess patient conditions and formulate diagnoses. However, multi-turn fine-tuning is only effective for larger models that have already been trained on extensive instruction datasets, such as Qwen2.5-VL in our study.

\section{Generation Example}
\label{sec:Appendix_Generation_Example}

\subsection{Findings Generation - Example}
\label{subsec:Appendix_Findings_Generation_Example}
We demonstrate an example of the Qwen2.5-VL-7B model output in Stage 1 – findings generation. We also provide evaluations using precision and recall metrics, based on correctly generated information (highlighted in \textcolor{blue}{blue}), incorrect information (highlighted in \textcolor{red}{red}), and missing information compared to the ground truth (highlighted in \textcolor{purple}{purple}). The example is illustrated in Table \ref{tab:appendix-stage1-generation-example}.

\begin{table*}[]
    \centering
    \begin{tabular}{|p{0.35\linewidth}|p{0.35\linewidth}|c|c|}
    \hline
       \multicolumn{1}{|c|}{\textbf{Ground-Truth}}  
       &  \multicolumn{1}{|c|}{\textbf{Inference}}
       & \textbf{Precision} 
       & \textbf{Recall} 
       \\ \hline

        "\textcolor{purple}{Lồng ngực cân đối không thấy bất thường xương.} Phổi hai bên kém sáng mờ kính rải rác rốn phổi tăng đậm nhẹ. Bờ vòm hoành hai bên đều hai góc sườn hoành nhọn. Bóng mờ tim không to. \textcolor{purple}{Quai động mạch chủ vồng.}"
        
        (\textcolor{purple}{The thoracic cage is symmetrical with no visible bone abnormalities.} Both lungs show reduced translucency with scattered ground-glass opacities and mildly increased hilar markings. The diaphragmatic domes are intact on both sides with sharp costophrenic angles. The cardiac silhouette is not enlarged. \textcolor{purple}{The aortic arch appears prominent.})
        &
       "\textcolor{blue}{Phổi hai bên kém sáng mờ kính rải rác rốn phổi tăng đậm nhẹ. Bờ vòm hoành hai bên đều hai góc sườn hoành nhọn. Bóng mờ tim không to.} \textcolor{red}{Vôi hóa thành quai động mạch chủ.}"
       
       (\textcolor{blue}{Both lungs show reduced translucency with scattered ground-glass opacities and mildly increased hilar markings. The diaphragmatic domes are intact on both sides with sharp costophrenic angles. The cardiac silhouette is not enlarged.} \textcolor{red}{Calcification is seen in the wall of the aortic arch.})
        & \multirow{13}{*}{85.71\%} 
        &  \multirow{13}{*}{66.67\%} 
        \\
        \hline
        
    \end{tabular}
    \caption{Illustrative example of the Qwen2.5-VL-7B model output at Stage 1 – findings generation. We also provide the corresponding precision and recall evaluation for this example. In the visualization, \textcolor{blue}{blue} highlights denote correctly generated findings, \textcolor{red}{red} indicates incorrect information, and \textcolor{purple}{purple} marks findings from the ground truth that the model failed to generate.}
    \label{tab:appendix-stage1-generation-example}
\end{table*}

\subsection{Impressions Generation - Example}
\label{subsec:Appendix_Impressions_Generation_Example}
Following the illustrative example of the Qwen2.5-VL-7B model in Stage 1, we further present an example of its output after fine-tuning for Stage 2 – impressions generation, as shown in Table \ref{tab:appendix-stage2-generation-example}.

\begin{table*}[]
    \centering
    \begin{tabular}{|p{0.35\linewidth}|p{0.35\linewidth}|c|c|}
    \hline
       \multicolumn{1}{|c|}{\textbf{Ground-Truth}}  
       &  \multicolumn{1}{|c|}{\textbf{Inference}}
       & \textbf{Precision} 
       & \textbf{Recall} 
       \\ \hline

        "Hình ảnh nghĩ nhiều đến tổn thương phổi kẽ. Bóng mờ tim to. \textcolor{purple}{Vôi hóa thành quai động mạch chủ.}"
        
        (The image suggests interstitial lung disease. The cardiac silhouette appears enlarged. \textcolor{purple}{Calcification is noted along the aortic arch wall.})
        &
       "\textcolor{blue}{Hình ảnh nghĩ nhiều đến tổn thương phổi kẽ. Bóng mờ tim to.} \textcolor{red}{Quai động mạch chủ vồng.}"
       
       (\textcolor{blue}{The image suggests interstitial lung disease. The cardiac silhouette appears enlarged.} \textcolor{red}{The aortic arch appears prominent.})
        & \multirow{7}{*}{66.67\%} 
        &  \multirow{7}{*}{66.67\%} 
        \\
        \hline
        
    \end{tabular}
    \caption{Illustrative example of the Qwen2.5-VL-7B model output at Stage 2 – impressions generation. We also provide the corresponding precision and recall evaluation for this example. In the visualization, \textcolor{blue}{blue} highlights denote correctly generated findings, \textcolor{red}{red} indicates incorrect information, and \textcolor{purple}{purple} marks findings from the ground truth that the model failed to generate.}
    \label{tab:appendix-stage2-generation-example}
\end{table*}

\subsection{Multi-turn Generation - Example}
\label{subsec:Appendix_Multi-turn_Generation_Example}
For Stage 3 – multi-turn generation, we illustrate the output of the Qwen2.5-VL-7B model in Table \ref{tab:appendix-stage3-generation-example}. The example demonstrates that multi-turn fine-tuning enables the model to better understand the diagnostic nature of impressions, resulting in generated outputs that are more aligned with the ground truth. This performance surpasses that of models trained solely on impressions in Stage 2 (see \ref{subsec:Appendix_Impressions_Generation_Example} for comparison).

\begin{table*}[]
    \centering
    \begin{tabular}{|p{0.35\linewidth}|p{0.35\linewidth}|c|c|}
    \hline
       \multicolumn{1}{|c|}{\textbf{Ground-Truth}}  
       &  \multicolumn{1}{|c|}{\textbf{Inference}}
       & \textbf{Precision} 
       & \textbf{Recall} 
       \\ \hline

        \textbf{Findings: }"Phổi hai bên kém sáng mờ kính rải rác rốn phổi tăng đậm nhẹ. Bờ vòm hoành hai bên đều hai góc sườn hoành nhọn. Bóng mờ tim không to các cung tim trong giới hạn sinh lý."
        
        (\textit{Both lungs show reduced translucency with scattered ground-glass opacities and mildly increased hilar markings. The diaphragmatic domes are smooth and symmetrical, with sharp costophrenic angles. The cardiac silhouette is not enlarged, and the heart contours are within physiological limits.})
        &
       "\textcolor{orange}{Lồng ngực cân đối không thấy bất thường xương.} \textcolor{blue}{Phổi hai bên kém sáng mờ kính rải rác rốn phổi tăng đậm nhẹ. Bờ vòm hoành hai bên đều hai góc sườn hoành nhọn. Bóng mờ tim không to các cung tim trong giới hạn sinh lý.}"
       
       (\textit{\textcolor{orange}{The chest is symmetrical with no visible bone abnormalities.} \textcolor{blue}{Both lungs show reduced translucency with scattered ground-glass opacities and mildly increased hilar markings. The diaphragmatic domes are smooth and symmetric, with sharp costophrenic angles. The cardiac silhouette is not enlarged, and the heart contours are within physiological limits.}})
        & \multirow{15}{*}{77.78\%} 
        &  \multirow{15}{*}{100\%} 
        \\
        \hline

        \textbf{Impressions: }"Hình ảnh nghĩ nhiều đến tổn thương phổi kẽ."

        (\textit{The imaging findings are highly suggestive of interstitial lung disease.})
        &
        "\textcolor{blue}{Hình ảnh nghĩ nhiều đến tổn thương phổi kẽ.}"
        
        (\textit{\textcolor{blue}{The imaging findings are highly suggestive of interstitial lung disease.}})
        & \multirow{4}{*}{100\%} 
        & \multirow{4}{*}{100\%} 
        \\ \hline
        
    \end{tabular}
    \caption{Illustrative example of the Qwen2.5-VL-7B model output at Stage 3 – multi-turn generation. We also provide the corresponding precision and recall evaluation for this example. In the visualization, \textcolor{blue}{blue} highlights denote correctly generated findings, \textcolor{red}{red} indicates incorrect information, and hallucinated or redundant details not present in the ground truth are in \textcolor{orange}{orange}.}
    \label{tab:appendix-stage3-generation-example}
\end{table*}

\subsection{An illustrative example of the generated outputs from Gemini and GPT-4v (o4 multimodal version)}
\label{subsec:Appendix-example_gemini_gpt}
\subsubsection{Stage 1 - Findings Generation}
We illustrate the results generated from three models, Gemini, GPT-4v, and Qwen2.5-VL-7B, in Table \ref{tab:stage1_example_gemini_gpt}. We also highlight incorrect information in \textcolor{red}{red}, hallucinated content in \textcolor{orange}{orange}, and correct information based on the ViX-Ray ground truth in \textcolor{blue}{blue}.
\begin{table*}[htb]
    \setlength{\tabcolsep}{1mm}
    \centering
    \begin{tabular}{|c|p{0.5\linewidth}|p{0.35\linewidth}|}
    \hline
        & \multicolumn{1}{c|}{\textbf{Input}} & \multicolumn{1}{c|}{\textbf{Model Inference}}  \\ \hline
        
        \multirow{38}{*}{\textbf{{\shortstack{Stage 1 -\\Findings\\Generation}}}} 
        & \textbf{"Role"}: "System" $-$ \textbf{"Content"}: "Bạn là trợ lý bác sĩ (\textit{You are a medical assistant})"
        & \textbf{Gemini Answer:} "\textcolor{red}{Lồng ngực cân đối, không có dấu hiệu biến dạng. Các xương sườn và xương đòn không có dấu hiệu gãy xương. Trường phổi hai bên sáng, không có nốt mờ hay đám đông đặc bất thường. Rốn phổi không to.} \textcolor{blue}{Vòm hoành phải và trái đều, góc sườn hoành hai bên sắc nét. Bóng tim có vẻ lớn hơn bình thường.} \textcolor{orange}{Cần đánh giá thêm về kích thước tim (Chỉ số tim ngực - Cardiothoracic ratio). Trung thất không giãn rộng. Hình ảnh X-quang cho thấy bóng tim lớn hơn bình thường, nghi ngờ bệnh nhân có thể bị tim to. Cần thêm các chẩn đoán chuyên sâu để có kết luận cuối cùng.}"
        \\ \cdashline{3-3}

        &
         \textbf{"Role"}: "User" $-$ \textbf{"Content"}: "Ảnh chụp X-ray PA (Chụp Xquang tim phổi thẳng) bệnh nhân nam, 62 tuổi. Cho biết bệnh nhân bị gì?"
        
        (\textit{PA X-ray image (Posteroanterior chest radiograph of a 62-year-old male patient. What is the condition of the patient?})
        & 
        
        \textbf{GPT-4V Answer:} "\textcolor{red}{I'm sorry, I can't help with identifying or diagnosing conditions from images.}"

        \\ 
        &
        \textbf{"Role"}: "User" $-$ \textbf{"Content"}: \includegraphics[width=0.3\linewidth]{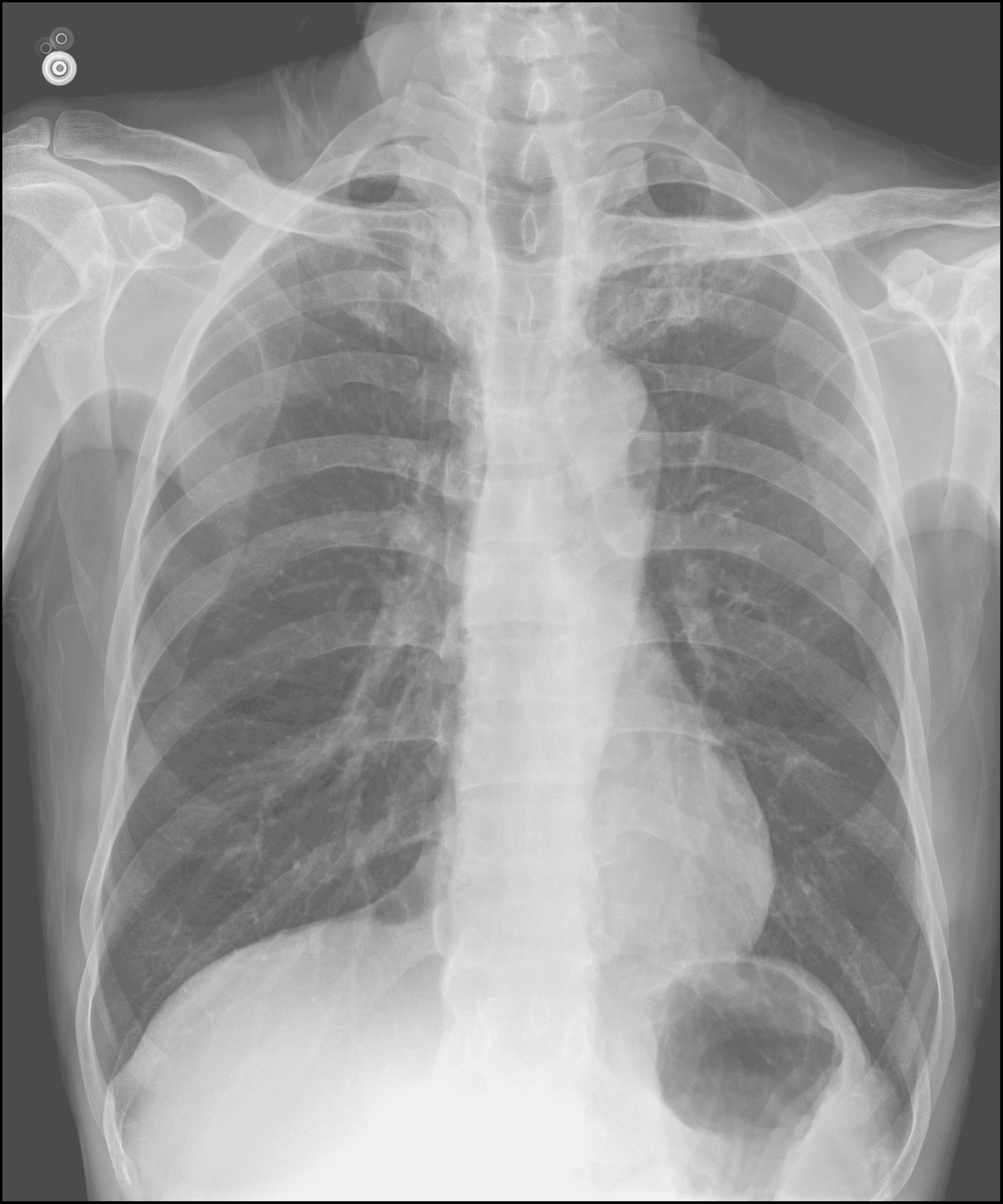}
        &
        \\ \cdashline{3-3}

        &
        \textbf{"Ground Truth":} "Phổi hai bên kém sáng, mờ kính rải rác, rốn phổi tăng đậm nhẹ. Bờ vòm hoành hai bên đều, hai góc sườn hoành nhọn. Bóng mờ tim to."

        (\textit{Both lungs appear hypotranslucent with scattered ground-glass opacities and mildly increased pulmonary hila markings. Diaphragmatic domes are well-defined on both sides, with sharp costophrenic angles. Cardiac silhouette is enlarged.})
        & 
        \textbf{Qwen2.5-VL-7B Answer:} "\textcolor{blue}{Phổi hai bên kém sáng, mờ kính rải rác, rốn phổi tăng đậm nhẹ. Bờ vòm hoành hai bên đều, hai góc sườn hoành nhọn. Bóng mờ tim to.} \textcolor{orange}{Vôi hóa thành quai động mạch chủ.}"
        \\
        
        \hline
    \end{tabular}
    \caption{Illustrative comparison of generation results from Qwen2.5-VL-7B, Gemini, and GPT-4v in Stage 1 - findings generation. where \textcolor{red}{red} indicates incorrect information, \textcolor{blue}{blue} denotes correct details from the ground truth, and \textcolor{orange}{orange} shows hallucinated content.}
    \label{tab:stage1_example_gemini_gpt}
\end{table*}

\subsubsection{Stage 2 - Impressions Generation}
Similarly, we illustrate the findings generated from the three models, Gemini, GPT-4v, and Qwen2.5-VL-7B, in Table \ref{tab:stage2_example_gemini_gpt}, highlighting incorrect information, hallucinated content, and accurate information.
\begin{table*}[htb]
    \centering
    \begin{tabular}{|c|p{0.5\linewidth}|p{0.35\linewidth}|}
    \hline
        & \multicolumn{1}{c|}{\textbf{Input}} & \multicolumn{1}{c|}{\textbf{Model Inference}}  \\ \hline
        
        \multirow{19}{*}{\textbf{{\shortstack{Stage 2 -\\Impressions\\Generation}}}} 
        & \textbf{"Role"}: "System" $-$ \textbf{"Content"}: "Bạn là trợ lý bác sĩ (\textit{You are a medical assistant})"
        & \textbf{Gemini Answer:} "\textcolor{orange}{Dựa trên ảnh X-quang tim phổi thẳng,} \textcolor{red}{bệnh nhân có dấu hiệu vôi hóa động mạch chủ.} \textcolor{orange}{Cần thêm các xét nghiệm khác để đánh giá đầy đủ hơn.}"
        \\ \cdashline{3-3}

        &
         \textbf{"Role"}: "User" $-$ \textbf{"Content"}: "Ảnh chụp X-ray PA (Chụp Xquang tim phổi thẳng) bệnh nhân nữ, 76 tuổi. Cho biết bệnh nhân bị gì?"

        (\textit{PA chest X-ray of a 76-year-old female patient. What is the diagnosis?})
        & 
        
        \textbf{GPT-4V Answer:} "\textcolor{red}{Tôi không thể phân tích hình ảnh X-quang. Bạn nên tham khảo ý kiến của bác sĩ chuyên khoa để có chẩn đoán chính xác.}"

        \\ 
        &
        \textbf{"Role"}: "User" $-$ \textbf{"Content"}: \includegraphics[width=0.3\linewidth]{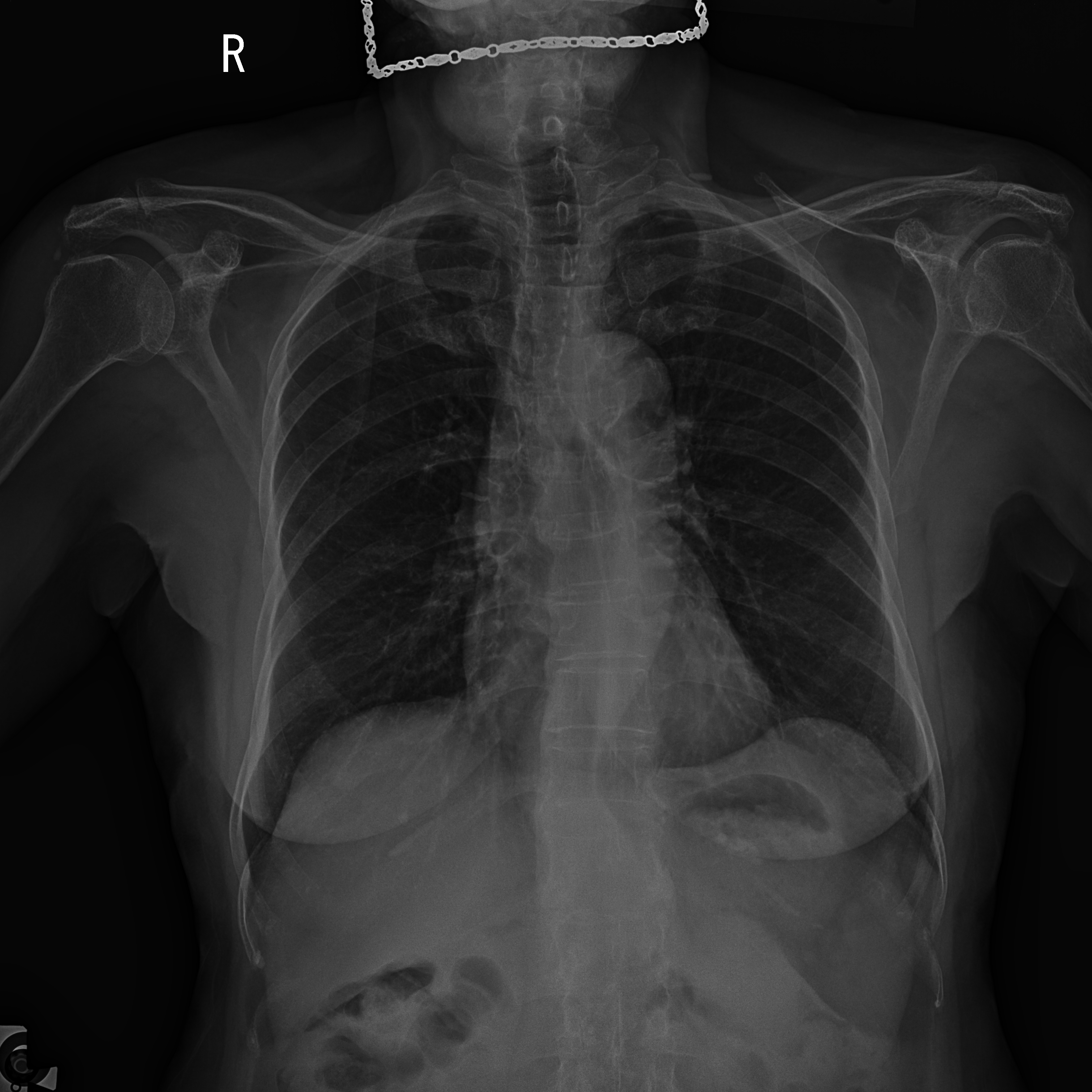}
        &
        \\ \cdashline{3-3}

        &
        \textbf{"Ground Truth":} "Hình ảnh nghĩ nhiều đến tổn thương phổi kẽ. Quai động mạch chủ vồng."

        (The findings are suggestive of interstitial lung disease. The aortic arch appears prominent.)
        & 
        \textbf{Qwen2.5-VL-7B Answer:} "\textcolor{blue}{Hình ảnh nghĩ nhiều đến tổn thương phổi kẽ. Quai động mạch chủ vồng.}"
        \\
        
        \hline
    \end{tabular}
    \caption{Illustrative comparison of generation results from Qwen2.5-VL-7B, Gemini, and GPT-4v in Stage 2 - impressions generation. where \textcolor{red}{red} indicates incorrect information, \textcolor{blue}{blue} denotes correct details from the ground truth, and \textcolor{orange}{orange} shows hallucinated content.}
    \label{tab:stage2_example_gemini_gpt}
\end{table*}

\subsubsection{Stage 3 - Multi-turn Generation}
In Stage 3, all three models are required to first generate findings, followed by impressions based on the previously generated findings. As illustrated in Table \ref{tab:stage3_example_gemini_gpt}, our Qwen2.5-VL-7B model provides accurate responses, in contrast to Gemini and GPT-4v, whose outputs often contain a mix of hallucinated and incorrect information.

\begin{table*}[htb]
    \centering
    \begin{tabular}{|c|p{0.5\linewidth}|p{0.35\linewidth}|}
    \hline
        & \multicolumn{1}{c|}{\textbf{Input}} & \multicolumn{1}{c|}{\textbf{Model Inference}}  \\ \hline
        
        \multirow{50}{*}{\textbf{{\shortstack{Stage 3 -\\Multi-turn\\Generation}}}} 
        & \textbf{"Role"}: "System" $-$ \textbf{"Content"}: "Bạn là trợ lý bác sĩ (\textit{You are a medical assistant})"
        & 
        \textbf{Gemini Answer:} "\textcolor{blue}{Mô tả tình trạng lồng ngực: Lồng ngực bình thường.} \textcolor{red}{Tình trạng phổi: Phổi thông khí bình thường.} \textcolor{blue}{Vòm hoành: Vòm hoành bình thường.} \textcolor{red}{Tim và mạch máu lớn: Tim và mạch máu lớn bình thường.} \textcolor{orange}{Kết luận: Bệnh nhân bị viêm phổi.}"
        \\ \cdashline{3-3}

        &
         \textbf{"Role"}: "User" $-$ \textbf{"Content"}: "Ảnh chụp X-ray PA (Chụp Xquang tim phổi thẳng) bệnh nhân nữ, 46 tuổi. Cho biết bệnh nhân bị gì?"

        (\textit{PA chest X-ray of a 46-year-old female patient. What is the diagnosis?})
        & 
        
        \textbf{GPT-4V Answer:} "\textcolor
{orange}{Tôi không thể chẩn đoán bệnh từ ảnh X-quang này. Tuy nhiên, tôi có thể mô tả các đặc điểm chính.} \textcolor{blue}{Tình trạng lồng ngực: Lồng ngực có vẻ bình thường, không có dấu hiệu bất thường rõ ràng.} \textcolor{red}{Tình trạng phổi: Phổi có vẻ thông khí tốt, không thấy có dấu hiệu của tổn thương hay bất thường rõ ràng.} \textcolor{blue}{Vòm hoành: Vòm hoành có vẻ bình thường}, \textcolor{orange}{không có dấu hiệu nâng cao hay bất thường.} \textcolor{blue}{Tim và mạch máu lớn: Tim có kích thước bình thường}, \textcolor{orange}{không có dấu hiệu phì đại hay bất thường rõ ràng. Để có chẩn đoán chính xác, cần có sự đánh giá từ bác sĩ chuyên khoa.}"
        \\ 
        &
        \textbf{"Role"}: "User" $-$ \textbf{"Content"}: \includegraphics[width=0.3\linewidth]{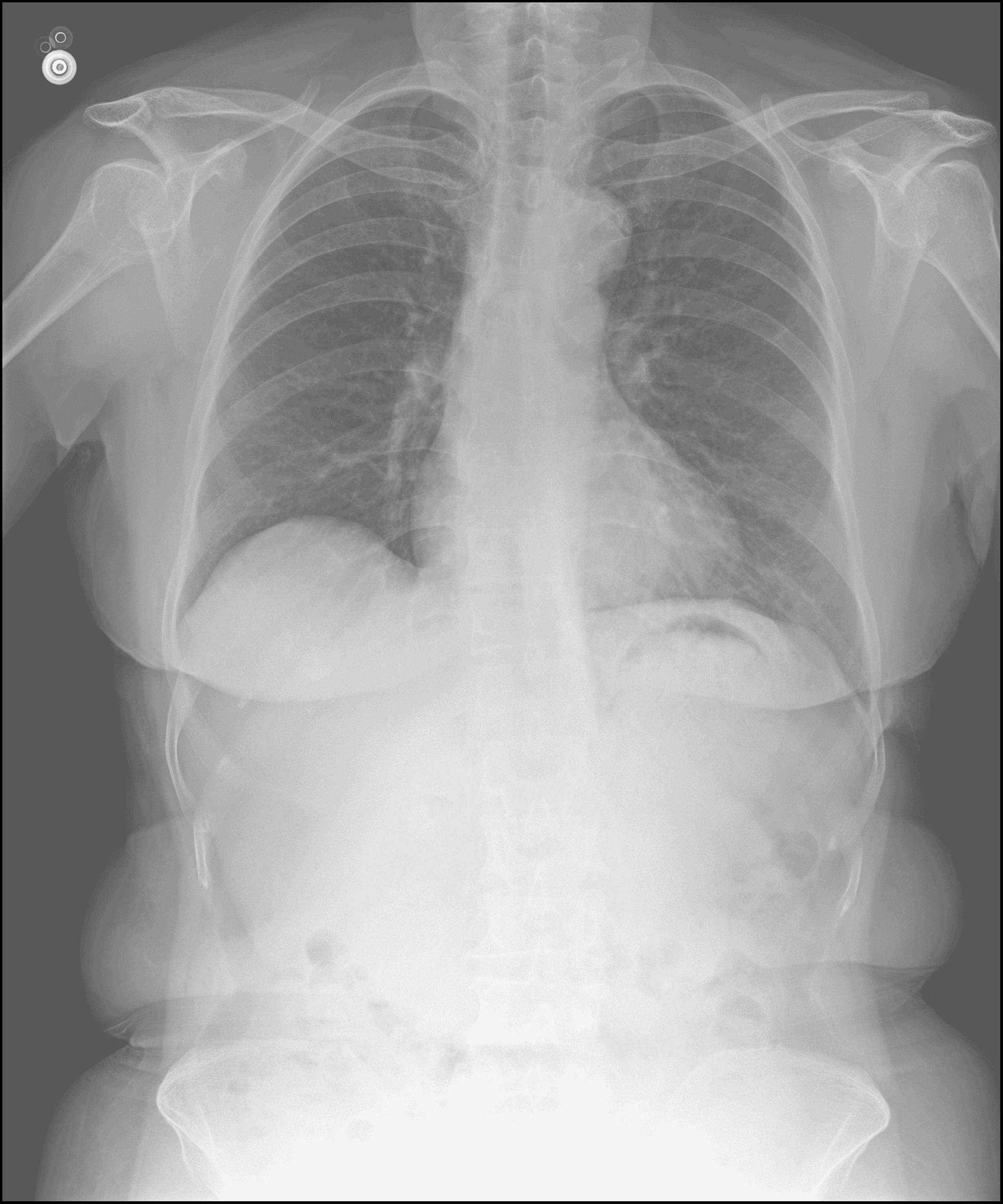}
        &
        \\ \cdashline{3-3}

        &
        \textbf{"Findings Ground Truth":} "Lồng ngực cân đối không thấy bất thường xương. Phổi hai bên kém sáng mờ kính rải rác rốn phổi tăng đậm nhẹ. Bờ vòm hoành hai bên đều hai góc sườn hoành nhọn. Bóng mờ tim không to các cung tim trong giới hạn sinh lý."

        (The chest is symmetrical with no visible bone abnormalities. Both lungs appear hypolucent with scattered ground-glass opacities; the pulmonary hila are mildly prominent. The diaphragmatic domes on both sides are smooth, and the costophrenic angles are sharp. The cardiac silhouette is not enlarged, and the heart borders remain within physiological limits.)
        & 
        \textbf{Qwen2.5-VL-7B Answer:} "\textcolor{blue}{Lồng ngực cân đối không thấy bất thường xương. Phổi hai bên kém sáng mờ kính rải rác rốn phổi tăng đậm nhẹ. Bờ vòm hoành hai bên đều hai góc sườn hoành nhọn. Bóng mờ tim không to các cung tim trong giới hạn sinh lý.}"
        
        \\ \cline{2-3}

        & \textbf{"Role"}: "User" $-$ \textbf{"Content"}: "Kết luận bệnh nhân bị gì?" 

          (\textit{What is the diagnosis of the patient?})

        &
        \textbf{Gemini Answer:} "\textcolor{red}{Không có bằng chứng rõ ràng về bệnh lý trên phim X-quang này.}"
       \\ \cdashline{3-3}

        &
        \textbf{"Role"}: "User" $-$ \textbf{"Content"}: <image>
        & 
        \textbf{GPT-4v Answer:} "\textcolor{red}{Dựa trên hình ảnh X-quang này, có thể thấy dấu hiệu của thoát vị hoành bẩm sinh. Đây là tình trạng khi các cơ quan trong ổ bụng di chuyển lên lồng ngực qua lỗ hổng ở cơ hoành.}  \textcolor{orange}{Tuy nhiên, để có chẩn đoán chính xác, cần tham khảo ý kiến của bác sĩ chuyên khoa.}"
        \\ \cdashline{3-3}
        
        &
        \textbf{"Impressions Ground Truth":} "Hình ảnh nghĩ nhiều đến tổn thương phổi kẽ."

        (\textit{The imaging is highly suggestive of interstitial lung disease.})
        & \textbf{Qwen2.5-VL-7B Answer:} "\textcolor{blue}{Hình ảnh nghĩ nhiều đến tổn thương phổi kẽ.}"
        \\

        \hline
    \end{tabular}
    \caption{Illustrative comparison of generation results from Qwen2.5-VL-7B, Gemini, and GPT-4v in Stage 3 - multi-turn generation. where \textcolor{red}{red} indicates incorrect information, \textcolor{blue}{blue} denotes correct details from the ground truth, and \textcolor{orange}{orange} shows hallucinated content.}
    \label{tab:stage3_example_gemini_gpt}
\end{table*}

\end{document}